\documentclass[review]{elsarticle}

\usepackage{lineno,hyperref}
\modulolinenumbers[5]

\usepackage{amsmath,amsthm,amssymb,epsfig,bm,amsmath}
 \usepackage{graphicx}
\usepackage{float}
\usepackage{csquotes}
\usepackage{amsmath}
\usepackage{mhchem}
\usepackage{xparse}
\usepackage{adjustbox}
\usepackage{MnSymbol}
\usepackage{mathdots}
\usepackage{makecell}
\usepackage{listings}
\usepackage{footnote}
\usepackage[flushleft]{threeparttable}
\usepackage{multirow}
\usepackage{relsize}
\usepackage{lettrine}
\usepackage{courier}
\usepackage{color} 
\definecolor{mygreen}{RGB}{28,172,0} 
\definecolor{mylilas}{RGB}{170,55,241}
\usepackage{amsmath,amsthm,amssymb,amsfonts} 
\usepackage{graphicx} 
\usepackage[margin=1in]{geometry} 
\usepackage[yyyymmdd]{datetime}
\usepackage{algorithm}
\usepackage[noend]{algpseudocode}
\usepackage{lipsum}
\usepackage{setspace}
\usepackage{siunitx}
\usepackage{pdfpages}
\usepackage{tikz}
\usetikzlibrary{positioning}
\usepackage{booktabs,caption}
\usepackage[]{hyperref}
\hypersetup{
 colorlinks  = true, 
 urlcolor   = blue, 
 linkcolor  = black, 
 citecolor  = blue 
}
\usepackage[numbers]{natbib}

\usepackage{listings}
\usepackage{color}
\definecolor{codegreen}{rgb}{0,0.6,0}
\definecolor{codegray}{rgb}{0.5,0.5,0.5}
\definecolor{codepurple}{rgb}{0.58,0,0.82}
\definecolor{backcolour}{rgb}{0.95,0.95,0.92}
\lstdefinestyle{mystyle}{
  backgroundcolor=\color{backcolour},  
  commentstyle=\color{codegreen},
  keywordstyle=\color{magenta},
  numberstyle=\tiny\color{codegray},
  stringstyle=\color{codepurple},
  basicstyle=\footnotesize,
  breakatwhitespace=false,     
  breaklines=true,         
  captionpos=b,          
  keepspaces=true,         
  numbers=left,          
  numbersep=5pt,         
  showspaces=false,        
  showstringspaces=false,
  showtabs=false,         
  tabsize=2,
  escapeinside={<@}{@>},
}
 
\usepackage{booktabs} 
\usepackage{siunitx} 

\usepackage{mathtools} 
\usepackage{gensymb} 
\usepackage{mhchem} 
\usepackage{fixltx2e} 
\usepackage{bbm}

\usepackage{multirow}

\usepackage{fancyhdr} 
\theoremstyle{definition}

\theoremstyle{definition}

\theoremstyle{remark}

\usepackage{soul} 
\soulregister\cite7
\soulregister\ref7
\soulregister\pageref7
\usepackage{bm}

\usepackage{framed} 

\usepackage{multicol} 

\usepackage{nomencl} 
\usepackage{tikz}
\makenomenclature
\usepackage[resetlabels,labeled]{multibib}

\renewcommand*\nompreamble{\begin{multicols}{2}}

\renewcommand*\nompostamble{\end{multicols}}

\usepackage{amssymb}
\usepackage{pifont}
\definecolor{light-gray}{gray}{0.95}

\usepackage[scaled]{helvet}
\usepackage[normalem]{ulem}

\newcommand{\rev}[1]{#1}                        

\makeatletter
\AtBeginDocument{\let\hl\@firstofone}
\makeatother

\journal{Journal to be decided}

\begin{document}


\begin{frontmatter}

\title{\large Physics-based Digital Twins for Integrated Thermal Energy Systems Using Active Learning}

\author{Umme Mahbuba Nabila$^{a}$, Paul Seurin$^{c}$, Linyu Lin$^{c}$, Majdi I. Radaideh$^{a,b,*}$}

\cortext[mycorrespondingauthor]{Corresponding Authors: Umme Nabila (unabila@umich.edu), Majdi I. Radaideh (radaideh@umich.edu)}

\address{$^{a}$Department of Nuclear Engineering and Radiological Sciences, University of Michigan, Ann Arbor, MI 48109, United States}

\address{$^{b}$Department of Computer Science and Engineering, University of Michigan, Ann Arbor, MI 48109, United States}

\address{$^{c}$Nuclear Science \& Technology Division, Idaho National Laboratory, Idaho Falls, ID 83415, United States}

\begin{abstract}

Real-time supervisory control of thermal energy distribution systems requires digital twins that are accurate, interpretable, and uncertainty-aware, yet remain data and computationally efficient. High-fidelity simulations alone are costly, while purely data-driven surrogates often lack robustness. To address these challenges, this work proposes an active learning (AL) framework that couples system-level Modelica simulations with four simpler physics-informed and data-driven surrogate modeling approaches: deterministic Sparse Identification of Nonlinear Dynamics with Control (SINDyC), its probabilistic multivariate-Gaussian extension (MvG-SINDyC), feedforward neural network (FNN), and gated recurrent unit (GRU) network. Tailored to each surrogate, model-specific AL query strategies are employed, including Mahalanobis-distance sampling in coefficient space for MvG-SINDyC and error-based sampling in prediction space for SINDyC, FNN, and GRU, allowing the learning process to prioritize dynamically informative trajectories.

The proposed approach is demonstrated on the glycol heat exchanger (GHX) subsystem of the Thermal Energy Distribution System (TEDS) at Idaho National Laboratory. Across key GHX outputs—the bypass mass flow rate $\dot{m}_{\mathrm{GHX}}$ and heat transfer rate $Q_{\mathrm{GHX}}$—the AL framework achieves comparable predictive accuracy using as few as one-fifth of the simulation trajectories required by random sampling. Among the evaluated surrogates, the GRU achieves the highest predictive fidelity, while SINDyC remains the most computationally efficient and interpretable. The probabilistic MvG-SINDyC surrogate further enables uncertainty quantification and exhibits the largest computational gains under AL. Overall, this work demonstrates that active learning provides a principled mechanism for identifying informative training regimes, enabling scalable, adaptive, and uncertainty-aware digital twins for real-time supervisory control of \rev{thermal energy distribution systems.}

\end{abstract}

\begin{keyword}
Digital Twin, Thermal Energy Distribution Systems, Sparse Identification of Nonlinear Dynamics with Control, Active Learning, Surrogate Modeling
\end{keyword}

\end{frontmatter}


\setstretch{1.3}

\section{Introduction}
\label{sec:intro}

\rev{Thermal energy distribution systems (TEDS) are complex thermal networks designed to store, transport, and deliver heat across interconnected components under time-varying operating conditions. They play a central role in advanced thermal energy infrastructures by coordinating the interaction of thermal energy storage (TES), heat exchangers, pumps, valves, piping loops, and distributed sensing and actuation systems. Through properly coordinated operation, TEDS can improve thermal flexibility, support load balancing, and enhance overall system efficiency} \cite{morton2020thermal,frick2020development}.

\rev{TEDS can also function as an enabling subsystem within broader integrated energy systems (IES), where thermal networks interface with generation assets, storage technologies, and end-use applications across multiple energy domains. In such settings, TEDS provides the thermal transport and management backbone needed to connect heat sources, storage units, and downstream processes in a flexible and coordinated manner. This role becomes especially important in nuclear-enabled IES architectures, where thermal energy storage and distribution can improve operational adaptability, support load following operation, and expand the range of viable non-electric applications} \cite{arvanitidis2023nuclear,bragg2020reimagining}. However, because these systems exhibit strongly coupled thermal-hydraulic behavior, nonlinear transients, and dynamic interactions across multiple subsystems, their accurate real-time monitoring, prediction, and supervisory control remain challenging.

Motivated by these challenges, recent research has increasingly emphasized the need for accurate, computationally efficient, and adaptive modeling frameworks for TEDS operation, control, and validation \cite{frick2020development,seurincontrol,el2024nuclear}. At Idaho National Laboratory (INL), a physics-based, system-level, dynamic Modelica model was developed to explore operating modes and component interactions of TEDS, thereby providing a benchmark for subsequent experiments \cite{frick2020development}. Follow-on work analyzed the operation and supervisory control of nuclear-coupled thermal energy storage systems using dynamic models \cite{mikkelson2022analysis} and this work was later extended to nuclear-renewable-hydrogen systems through hierarchical Modelica architectures \cite{jacob2023modeling}. Complementary studies using OpenModelica demonstrated hybrid desalination concepts that integrate TES and improve grid flexibility \cite{hills2021dynamic}. Beyond INL, modular and object-oriented modeling frameworks (e.g., Modelica) have enabled plug-and-play configurations that couple nuclear reactors with hydrogen production processes and thermal energy storage, allowing flexible system design and scenario testing under realistic operating conditions \cite{masotti2023modeling}. At the planning scale, technology-rich optimization frameworks such as the TIMES-Europe model integrate power, industrial, and transport sectors across the EU-27+UK, yielding policy-relevant decarbonization pathways that complement component-level dynamic analyses \cite{luxembourg2025times}. At the system level, optimization tools have been cross-validated with dynamic simulations for islanded microgrids that involve molten-salt TES \cite{williams2024modeling}, while real-time, hardware-in-the-loop experiments evaluated power-quality and control issues regarding nuclear-renewable-hydrogen integration \cite{gautam2025digital}. Existing optimization and planning frameworks are primarily designed to inform long-term system design, capacity expansion, and market participation, typically operating on hourly to annual time scales. While these tools provide valuable strategic insights, they are not intended for real-time operation or closed-loop control. However, a gap remains between these planning- and design-oriented frameworks and the requirements of real-time system operation and control. Digital twins (DTs) address this gap by providing high-resolution, dynamic representations of physical systems that can be continuously updated using operational data. Rather than replacing planning or design-oriented optimization frameworks, digital twins complement them by translating high-level decisions into actionable, real-time operational guidance.

Traditionally, TEDS has been operated using decentralized proportional-integral controllers. However, such control strategies fails to capture cross-component coupling and system constraints. To overcome these limitations, supervisory controllers are being developed to coordinate multi-component signals and optimize system-wide performance \cite{lin2024development}. These strategies improve efficiency and responsiveness, offering faster, more reliable control than that afforded by manual operation \cite{lin2024autonomous}. Among various advanced control methods, model predictive control (MPC) is particularly well suited for TEDS. MPC uses a simplified model of the system to predict how the system will evolve over a short time horizon and evaluates the effect of different control actions. Based on these predictions, it selects control inputs that guide the system toward desired operating conditions while respecting operational constraints \cite{kouvaritakis2016model}. At each control step, only the first optimized input is applied, and the optimization is repeated using updated measurements. This receding-horizon strategy enables adaptive, real-time control under uncertainty, making MPC especially suitable for complex, multi-input and multi-output systems such as TEDS \cite{seurin2026uncertainty}. Although detailed physics-based TEDS models provide high accuracy, their computational cost—arising from numerical stiffness and solver latency— limits their direct applicability in real-time MPC applications \cite{frick2021validation}. This limitation motivates the use of surrogate models that approximate system dynamics using data from high-fidelity simulations, enabling substantially faster prediction while retaining sufficient accuracy for control.

Among different data-driven approaches available, physics-informed Sparse Identification of Nonlinear Dynamics with Control (SINDyC) has gained attention for its simplicity, interpretability, and ability to recover governing equations directly from data \cite{brunton2016discovering, kaiser2018sparse}. Because SINDyC produces analytical state–space representations, the resulting models offer physical insight and can be embedded directly within optimization-based control frameworks. Their differentiable structure further enables efficient gradient-based optimization, making SINDyC particularly attractive for model predictive control (MPC) applications. However, robust control requires not only speed and interpretability, but also reliable uncertainty quantification (UQ). To address this requirement, Seurin et al. introduced a probabilistic extension of SINDyC designed specifically to support uncertainty-aware control, aggregating thousands of deterministic SINDyC realizations into a single probabilistic surrogate model --- multivariate Gaussian SINDyC (MvG-SINDyC) \cite{seurin2026uncertainty}. However, in this implementation, trajectory selection relied on random sampling, an approach that does not explicitly account for the relative information content of individual datasets. Following the PySINDy convention, each simulation dataset is referred to herein as a trajectory \cite{kaptanoglu2021pysindy}.

Building on the probabilistic MvG-SINDyC foundation, \rev{we introduced an active learning (AL) strategy to achieve data-efficient surrogate modeling for TEDS.} Rather than relying on random or exhaustive sampling, the proposed AL framework adaptively identifies the most informative simulation trajectories using two complementary query strategies: 
(1) Mahalanobis-distance sampling in the coefficient space for MvG-SINDyC, and 
(2) Error-reduction sampling in the prediction space for other surrogates. 
The Mahalanobis distance quantifies deviations between model coefficients and the experimental baseline while accounting for both scale and correlation \cite{de2000mahalanobis}. This targeted selection accelerates convergence toward accurate surrogates using fewer simulations, thereby enabling scalable digital twins for MPC without compromising robustness or interpretability.

While physics-informed surrogates such as SINDyC provide interpretability and computational efficiency, they can be limited in capturing nonlinear and temporal dynamics when key states are unobserved or only partially measurable. 
In contrast, machine-learning-based models, such as feedforward neural networks (FNNs) and gated recurrent units (GRUs), offer strong predictive capabilities for nonlinear and sequential systems \cite{lecun2015deep, goodfellow2016deep, chung2014empirical}, and have been successfully applied as surrogate models in a range of energy-system applications \cite{radaideh2020neural,radaideh2019combining,saleem2020application}. 
However, these black-box models typically require large datasets to generalize reliably, which poses challenges in data-limited settings involving costly experiments or computationally intensive simulations, and offers limited interpretability. This study therefore extends the active learning (AL) framework beyond SINDyC to include feedforward neural network (FNN) and gated recurrent unit (GRU) surrogates, enabling a unified assessment of equation-based modeling, probabilistic modeling, and black-box modeling approaches. The resulting comparison highlights fundamental trade-offs among model transparency, uncertainty representation, and prediction performance, and demonstrates how targeted AL-driven sampling improves learning efficiency, reliability, and robustness across surrogate classes.

More broadly, this work advances surrogate modeling strategies that integrate physics-consistent structure with the flexibility of machine learning, supporting the development of reliable and adaptive digital twin methodologies for complex thermal-hydraulic systems. This comparative study contributes to the advancement of DTs for TEDS in three key ways:

\begin{enumerate}
    \item It demonstrates that AL can substantially accelerate surrogate-model convergence across diverse modeling approaches---including interpretable physics-informed surrogates (MvG-SINDyC) and black-box neural networks (FNN and GRU)---thereby improving both data efficiency and predictive accuracy.
    \item It clarifies the trade-offs among interpretability, UQ, and predictive fidelity, offering a framework for selecting or hybridizing surrogates based on the operational requirements of DTs and MPC controllers.
    \item It validates the trained surrogate models by comparing their predictions against experimental measurements.
\end{enumerate}

The remainder of this paper is organized as follows. Section \ref{sec:rel} reviews related work on surrogate modeling, digital twins, and data-driven control methodologies relevant to this study. Section \ref{sec:data} describes the experimental data and simulation models used for model training and validation. Section \ref{sec:method} details the theoretical foundations and methodologies employed for surrogate model construction, including SINDyC, MvG-SINDyC, FNN, and GRU, and introduces the conceptual framework for applying AL to DT development. Section \ref{sec:results} presents a comparative analysis of AL versus random sampling across all the modeling approaches explored, and discusses the implications of these findings for for scalable and reliable DTs for supervisory control. Finally, Section \ref{sec:conc} summarizes the key contributions and future research directions.

\begin{table*}[!t]  
\small
\begin{framed}
\begin{footnotesize}
\nomenclature{SINDyC}{Sparse Identification of Nonlinear Dynamics with Control}
\nomenclature{MvG-SINDyC}{multivariate-Gaussian SINDyC}
\nomenclature{FNN}{feedforward neural network}
\nomenclature{GRU}{gated recurrent unit}
\nomenclature{AL}{active learning}
\nomenclature{ML}{machine learning}

\nomenclature{MPC}{model predictive control}
\nomenclature{DT}{digital twin}
\nomenclature{TEDS}{thermal energy distribution system}
\nomenclature{TES}{thermal energy storage}
\nomenclature{GHX}{glycol heat exchanger}

\nomenclature{RMSE}{root mean squared error}
\nomenclature{UQ}{uncertainty quantification}

\end{footnotesize}

\printnomenclature

\end{framed}

\end{table*}

\section{Related Work}
\label{sec:rel}

\rev{Recent research relevant to this study spans surrogate modeling, reinforcement learning (RL), digital twins, and fault detection for dynamic thermal energy systems. In the area of surrogate modeling, deep Gaussian processes have been developed to emulate highly nonlinear nuclear simulations with improved uncertainty characterization} \cite{radaideh2020surrogate}, and surrogate-driven variance-based sensitivity analysis has been applied to thermal storage tank in TEDS\cite{sene2025surrogate}. Additional reduced-order techniques such as simplified matching pursuits enable fast reconstruction of three-dimensional temperature fields from sparse sensor measurements \cite{price2024simplified}. Related surrogate modeling studies have also explored the use of machine-learning surrogates for the EnergyPLAN framework to speed up country-level low-carbon energy system optimization \cite{prina2024machine}, as well as systematic studies of surrogate model architecture, sampling, and scaling choices for multi-energy system design \cite{ledee2025improved}. \rev{Data-driven distributionally robust formulations further leverage historical data to construct ambiguity sets for stochastic scheduling and dispatch under uncertainty} \cite{li2023data,zhou2025data}.

\rev{Reinforcement learning and advanced optimization have also seen rapid progress in control-oriented engineering and energy applications.Physics-informed RL has been applied to optimize nuclear assembly design under various reactor-physics constraints} \cite{radaideh2021physics}, \rev{while prioritized experience replay mechanisms accelerate hybrid evolutionary--swarm search algorithms for nuclear fuel optimization} \cite{radaideh2022pesa}. \rev{These developments culminated in an optimization framework, which unifies evolutionary algorithms, neuroevolution, and RL for scalable optimization of carbon-free energy systems} \cite{radaideh2023neorl}.\rev{In parallel, deep RL has been applied to optimal dispatch in coupled electricity--heat--gas systems with thermal storage, heat loads, and cross-domain conversion devices} \cite{zhang2023hybrid}, \rev{as well as to load-following control and criticality search for nuclear microreactors, thereby providing useful algorithmic context for supervisory decision-making under uncertainty} \cite{tunkle2025nuclear,radaideh2025multistep}. \rev{These data-driven optimization and RL-based control strategies offer useful methodological foundations for supervisory control, scheduling, and adaptive decision making in thermal energy distribution systems.}

\rev{Digital-twin research has further advanced through variational formulations that integrate latent-space inference with physics-based priors for real-time state estimation and uncertainty tracking} \citep{burnett2025variational}, \rev{complemented by fast surrogate-based reconstruction and reduced-order modeling approaches that support online monitoring and control. In parallel, significant progress has been made in fault detection and predictive diagnostics for safety-critical energy components. Neural time-series forecasting has been used to predict the evolution of loss-of-coolant accidents from plant signals} \citep{radaideh2020neural}, \rev{while convolutional and feedforward neural networks have been demonstrated for fault detection in accelerator power systems} \citep{radaideh2022application}. \rev{Recurrent and ConvLSTM autoencoder models enable high-fidelity anomaly detection in power-electronics signals} \citep{radaideh2022time}, and ensemble learning approaches further improve robustness across diverse fault types and operating regimes \citep{radaideh2023early}. Moreover, multi-module conditional variational autoencoders have been applied to accurately predict high-voltage converter modulator faults in particle accelerators \citep{alanazi2023multi}.\rev{Together, these studies establish a methodological foundation for combining surrogate-assisted modeling, adaptive learning, and physics-informed inference in digital twins for dynamic thermal systems.}


\rev{Collectively, prior works have established a strong foundation for modeling, simulation, optimization, and experimental validation which is valuable for the present problem setting. However, relatively limited work has focused on active learning-enabled digital twin development in settings where strongly coupled thermal-hydraulic behavior, supervisory control relevance, uncertainty awareness, and data efficiency must all be addressed simultaneously.} While high-fidelity optimization and planning frameworks offer valuable insights into long-term system design and operation, these approaches are either computationally restrictive for real-time control or lack explicit treatment of uncertainty and data efficiency. Existing surrogate modeling efforts have made progress toward faster prediction, yet challenges remain in balancing interpretability, uncertainty quantification, and predictive accuracy under limited data availability. These gaps motivate the development of data-efficient, uncertainty-aware surrogate modeling approaches that can be embedded within digital twin and MPC frameworks, which is the focus of the present work.

\section{Experiment and Simulation Data Collection} \label{sec:data}

\subsection{TEDS Experimental Facility} \label{sec:teds}

\rev{TEDS, an experimental testbed within INL's Dynamic Energy Transport and Integration Laboratory, was designed to investigate the dynamics of coupled heat transfer, energy storage, and distribution components under realistic operating conditions}\cite{lin2024development}. It serves as a controllable, instrumented platform for validating reduced-order models, developing supervisory control strategies, and benchmarking DT methodologies.

\begin{figure}[htbp]
    \centering
    \begin{minipage}[t]{0.48\textwidth}
        \centering
        \includegraphics[width=\linewidth]{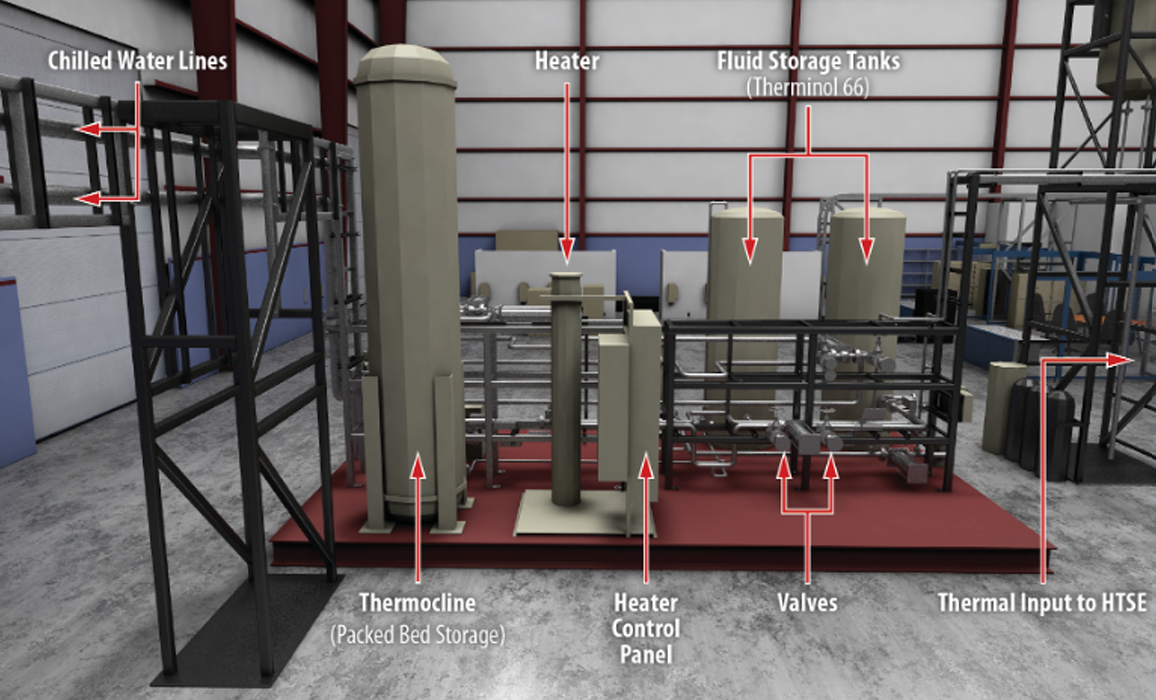}
        \caption{Experimental TEDS facility, showing the heater, TES tank, thermocline, fluid storage tanks, and control valves \cite{morton2020thermal}.}
        \label{fig:teds_photo}
    \end{minipage}%
    \hfill
    \begin{minipage}[t]{0.48\textwidth}
        \centering
        \includegraphics[width=\linewidth]{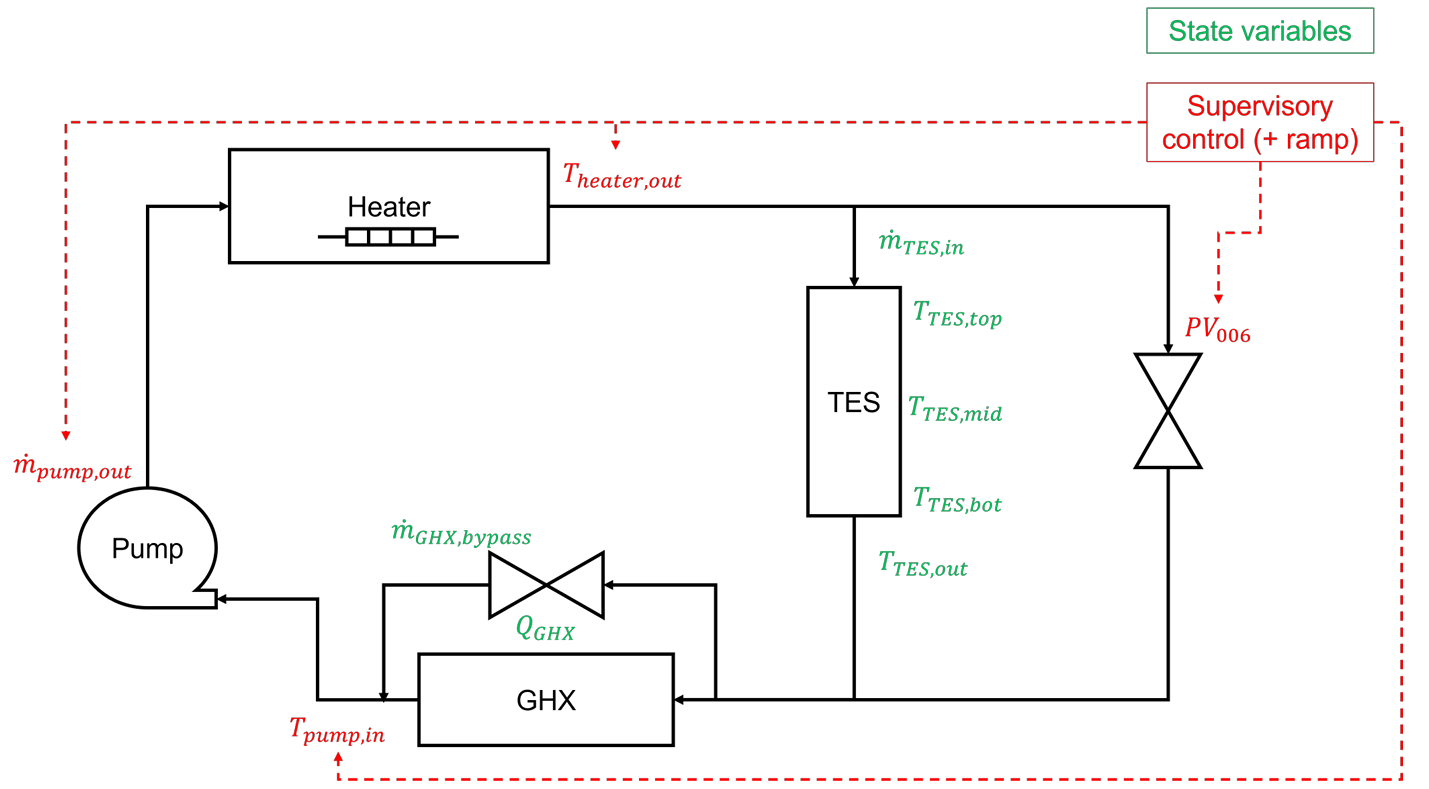} 
        \caption{Simplified schematic of the TEDS loop and supervisory control architecture \cite{seurin2026uncertainty}.}
        \label{fig:teds_schematic}
    \end{minipage}
\end{figure}

Figure~\ref{fig:teds_photo} shows a schematic of the TEDS facility, illustrating the heater, TES tank, thermocline, fluid storage tanks, and control valves \cite{morton2020thermal}. TEDS is comprised of a 200~kW Chromalox electrical heater, a single-tank packed-bed TES, and a glycol heat exchanger (GHX), along with the associated piping, five control valves, and distributed temperature and flow sensors. TES consists of a vertical cylindrical tank filled with high-conductivity alumina beads (0.125~in.) that act as a filler medium, reducing the required fluid inventory and enhancing thermal stratification. Charging occurs when hot Therminol-66 is introduced at the top of the tank while cooler fluid is withdrawn from the bottom, gradually forming a thermocline. During discharging, the flow is reversed, with colder fluid being sent to the bottom and warmer fluid being extracted from the top, enabling efficient energy release. The GHX, constructed as a shell-and-tube heat exchanger, transfers heat between the Therminol loop and a secondary ethylene glycol loop, which emulates industrial or grid-connected demand. The Chromalox heater supplies high-temperature fluid to initiate charging cycles and regulate operating conditions.

The experiment cycles through five primary operating phases: startup (warm-up), charging, standby, discharging, and cooldown. For modeling and control, TES state vector includes the inlet mass flow rate $\dot m_{\mathrm{tes,in}}$, outlet temperature $T_{\mathrm{tes,out}}$, and three internal node temperatures $T_{\mathrm{top}},\, T_{\mathrm{mid}},\, T_{\mathrm{bot}}$ that resolve the thermocline. The GHX state vector is represented by the bypass flow $\dot m_{\mathrm{GHX}}$ and the extracted heat rate $Q_{\mathrm{GHX}}$. From a physical actuation perspective, the system is directly manipulated through these actuators: the valve position $PV_{006}$, the pump outlet mass flow rate $\dot m_{\mathrm{pump,out}}$, and the heater outlet temperature $T_{\mathrm{heater,out}}$. The pump inlet temperature $T_{\mathrm{pump,in}}$ is not directly actuated; instead, it is a measured variable determined by upstream thermal mixing, TES outlet conditions, and the flow path selected by $PV_{006}$.
For control-oriented modeling and formulation, $T_{\mathrm{pump,in}}$ is included as an input variable alongside $PV_{006}$, $\dot m_{\mathrm{pump,out}}$, and $T_{\mathrm{heater,out}}$ to account for its influence on downstream dynamics. Together, these control inputs and measured states govern the coupled TES–GHX behavior analyzed in this study.

The present study emphasizes the discharging process (approximately 9,180--14,640~seconds of the experiment process) during which the packed-bed TES delivers stored heat to the GHX loop. The supervisory controller actuates $PV_{006}$ (located on the right side of the TES module) and the circulation pump to route fluid from the TES through the GHX, as illustrated in Figure~\ref{fig:teds_schematic}. The discharge behavior (Experiment) can be observed in Figure~\ref{fig:GHX}. In these plots, $PV_{006}=1$ indicates that the valve is open to the GHX path, while $PV_{006}=0$ indicates that it is closed and that the flow is redirected through the bypass line. When $PV_{006}$ is open, hot fluid from the upper TES layer flows through the GHX, transferring heat to the glycol loop. As a result, the GHX heat rate $Q_{\mathrm{GHX}}$ increases rapidly, while the bypass flow $\dot{m}_{\mathrm{GHX}}$ decreases toward zero. The pump maintains an almost constant outlet flow, ensuring stable loop circulation. Over time, TES gradually cools from the top downward as the hot region (thermocline) moves lower in the bed. This reduces the temperature difference driving the heat transfer, leading to a slow decline in $Q_{\mathrm{GHX}}$. At around 2,000~seconds, $PV_{006}$ switches from open to closed ($1 \rightarrow 0$). The flow is then redirected through the bypass, causing $\dot{m}_{\mathrm{GHX}}$ to rise sharply while $Q_{\mathrm{GHX}}$ drops nearly to zero. The present analysis specifically considers the GHX states $\dot m_{\mathrm{GHX}}$ and $Q_{\mathrm{GHX}}$, which are critical for characterizing TEDS's transient heat delivery capability and for evaluating surrogate model predictions. This configuration allows simultaneous exploration of trade-offs between model fidelity, interpretability, and computational tractability---all while being anchored to experimental dataset.

\subsection{Modelica Simulation Framework} \label{sec:modelica}

\begin{figure}[htbp]
    \centering
    \begin{minipage}[t]{0.48\textwidth}
        \centering
        \includegraphics[width=\linewidth]{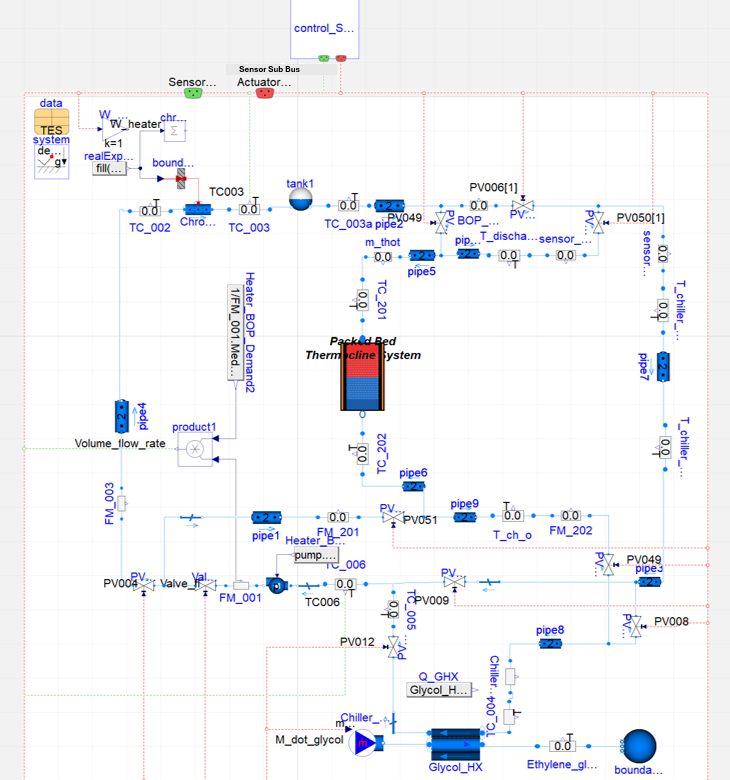}
        \caption{TEDS Modelica framework.}
        \label{fig:mod_teds}
    \end{minipage}%
    \hfill
    \begin{minipage}[t]{0.48\textwidth}
        \centering
        \includegraphics[width=\linewidth]{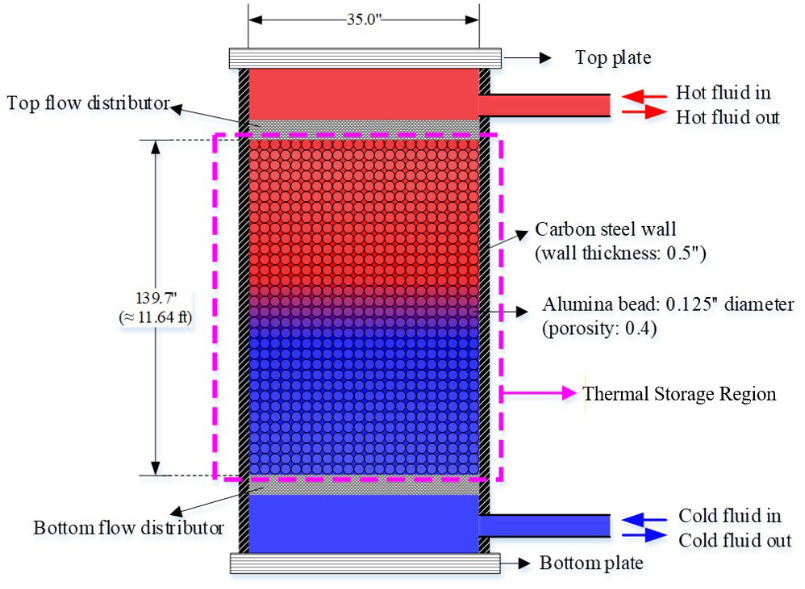} 
        \caption{Thermocline Energy Storage Tank \cite{qin2022thermal}}.
        \label{fig:tank}
    \end{minipage}
\end{figure}

While the TEDS experiment provided measurements of coupled TES-GHX dynamics, only a single discharge trajectory was available for each quantity of interest. To address this data limitation and enable robust surrogate training, this study leverages an existing physics-based Modelica representation of TEDS. The model employs an equation-based formulation that enforces conservation of mass and energy across all components and is executed using the commercial platform Dymola. The overall architecture and component-level implementations follow the validated TEDS Modelica framework developed at Idaho National Laboratory and documented in \cite{frick2020development}. No modifications were made to the underlying Modelica formulation; the model was used solely as a trajectory-generation engine. As shown in Figure~\ref{fig:mod_teds}, the simulation environment mirrors the experimental configuration, including the TES packed bed and the GHX loop, valves, heater, and pump, thus providing a virtual platform for controlled trajectory generation. In particular, the thermocline thermal energy storage (TES) unit, shown in Figure~\ref{fig:tank}, is modeled as a one-dimensional packed-bed system governed by a modified form of the classical Schumann equations for fluid flow through porous media \cite{frick2021validation}. The thermocline consists of a circulating fluid phase and a stationary filler (solid) phase, that exchange thermal energy through interphase convection. Separate energy balances are formulated for the fluid and filler phases and are coupled via an interphase convective heat-transfer term. The interphase convective heat-transfer coefficient is evaluated using an empirical correlation for flow through porous media.

The transient energy balance for the fluid phase over an axial control volume is written as,

\begin{equation}
\rho_f C_f \varepsilon \pi R^2
\left(
\frac{\partial T_f}{\partial t}
+ U \frac{\partial T_f}{\partial z}
\right)
=
h_c S_r \left( T_r - T_f \right)
+ \dot{Q}_{\mathrm{loss}},
\label{eq:fluid_energy}
\end{equation}

where $\rho_f$ and $C_f$ are the density and specific heat capacity of the fluid material, $T_f$ and $T_r$ are the fluid and filler temperatures, respectively, $\varepsilon$ is the bed porosity, $h_c$ is the convective heat-transfer coefficient, $R$ is the thermocline tank radius, and $\dot{Q}_{\mathrm{loss}}$ represents heat losses through the tank wall. The superficial axial velocity of the fluid is defined as, 
\begin{equation}
U = \frac{\dot{m}}{\rho_f \varepsilon \pi R^2},
\label{eq:velocity}
\end{equation}
The effective heat-transfer surface area of the filler material per unit axial length is given by,
\begin{equation}
S_r = \frac{f_s \pi R^2 (1-\varepsilon)}{r_{\mathrm{fill}}},
\label{eq:surface_area}
\end{equation}
where $f_s$ is a shape factor and $r_{\mathrm{fill}}$ is the characteristic radius of the filler particles. Axial conduction within the fluid is neglected. 

The corresponding transient energy balance for the filler material is given by,

\begin{equation}
h_c S_r \left( T_r - T_f \right)\, dz
=
- \rho_r C_r (1-\varepsilon)\,\pi R^2 \, dz \,
\frac{\partial T_r}{\partial t},
\label{eq:filler_energy}
\end{equation}

where $\rho_r$ and $C_r$ are the density and specific heat capacity of the filler material. This equation accounts for thermal energy storage within the filler due to interphase heat exchange with the fluid. 


To enrich the training set, synthetic actuator trajectories were generated around the experimental operating points. Specifically, the ranges of the four actuators---valve position $PV_{006}$, pump outlet flow $\dot m_{\mathrm{pump,out}}$, pump inlet temperature $T_{\mathrm{pump,in}}$, and heater outlet temperature $T_{\mathrm{heater,out}}$---were expanded beyond the nominal discharge setpoints. A Sobol sequence sampler was then employed to draw 374 valid candidate time series across this expanded operating envelope. Sobol sampling determines which actuator trajectories are simulated. This quasi-random sampling ensured good coverage of the high-dimensional control space while also avoiding clustering effects typical of purely random draws. There is no exact one-to-one match between the experimental actuation signals and any single Modelica simulation. During the experiments, multiple actuator commands are applied simultaneously. In contrast, to ensure numerical stability of the Modelica solver, actuation events in the simulations are intentionally staggered in time (typically by approximately 50s) so that each actuation occurs separately. As a result, the timing of simulated actuation signals does not coincide exactly with the experimental actuation times. Numerical sensitivity analyses reported in prior Modelica-based studies have shown that these small timing offsets do not significantly affect the magnitudes of the quantities of interest. In addition, several actuator-related variables such as temperature are not directly controlled but achieved by adjusting heater or pressure drop in the experimental setup. Because these variables are governed by internal control dynamics rather than direct actuation, discrepancies between experimental and simulated trajectories are expected.

\begin{figure}[!h]
    \centering
    \includegraphics[scale=0.55]{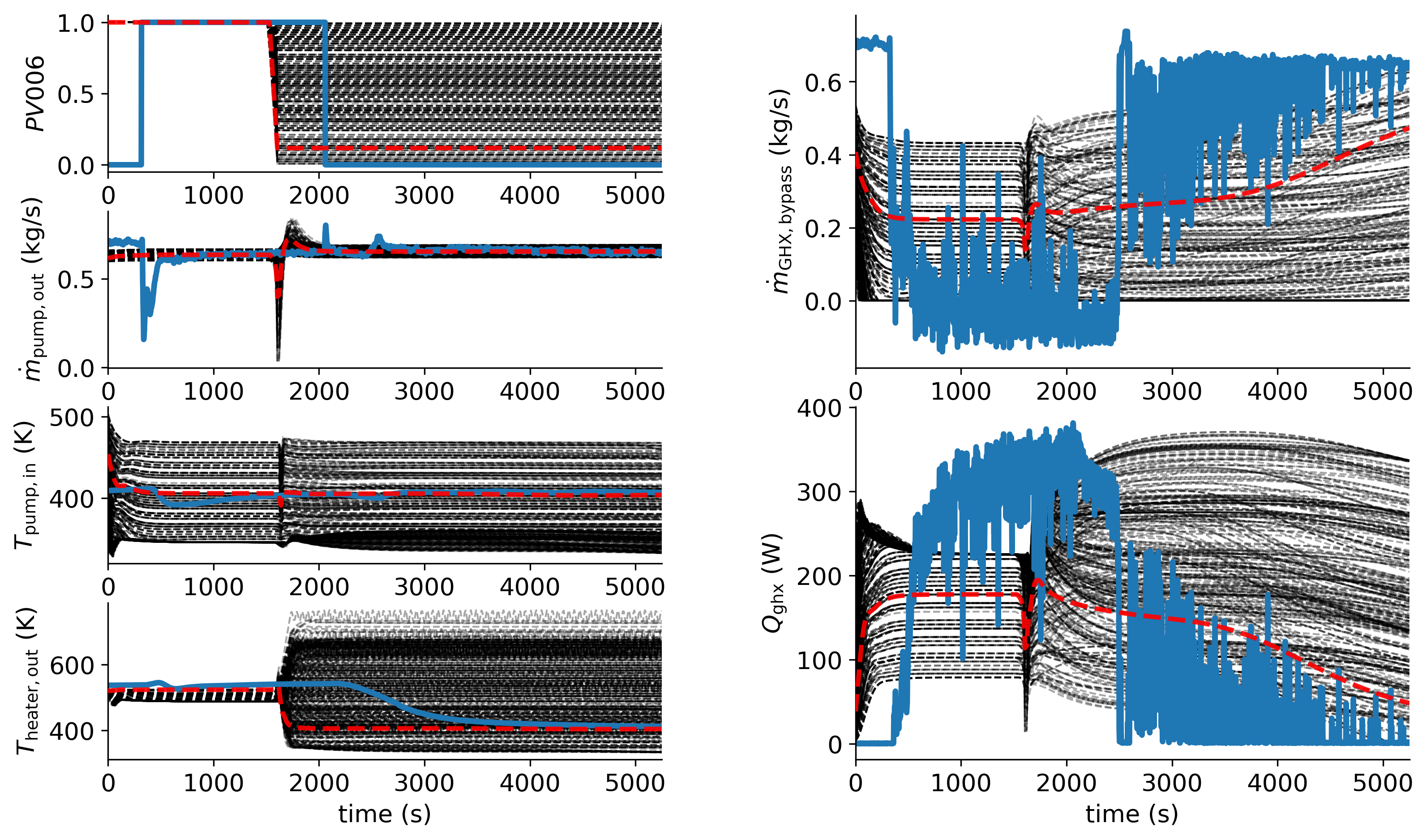}
    \caption{Experimental and Modelica actuators and states of GHX. The blue lines represent the experimental data, the black curves represent different Modelica simulations, and red curve denotes the Modelica run (Run 117) whose actuator inputs best match the experiment.}
    \label{fig:GHX}
\end{figure}

Each actuator signal was refined to a common resolution of 5,251 points over a simulation window of 5,460 seconds, to ensure consistency with the experimental discharge duration. This corresponds to an effective sampling interval of approximately 1.04 s, or a sampling frequency of about 0.96 Hz. This resampling was performed to ensure consistency across all signals and to align the simulation data with the duration of the experimental discharge. The experimental and simulated trajectories are compared in Figure~\ref{fig:GHX}, with the black curves denoting the 374 valid Modelica-generated signals. The red dashed curves indicate the single Modelica trajectory selected as the closest match to the experimental actuation profile, and the blue curves representing the de-noised experimental trajectories, filtered via a Savitzky-Golay filter. This filter is applied to reduce measurement noise while preserving trends and derivatives. The figure highlights how the synthetic trajectories explore a broader range of dynamics while still remaining consistent with the experimental observations.

In addition, a direct quantitative comparison was performed between the experimental trajectory and the Modelica simulation whose actuation inputs most closely match the experiment. Under these matched conditions, the Modelica model predicts the GHX bypass mass flow with an RMSE of 0.27 kg/s and the GHX heat transfer rate with an RMSE of 115 W. While the Modelica model captures the overall trends, these results indicate that non-negligible discrepancies remain under experimental conditions.

\section{Methodology} 
\label{sec:method}

\subsection{Digital Twin Framework} \label{sec:dt}
\begin{figure}[!h]
    \centering
    \includegraphics[width=0.9\textwidth]{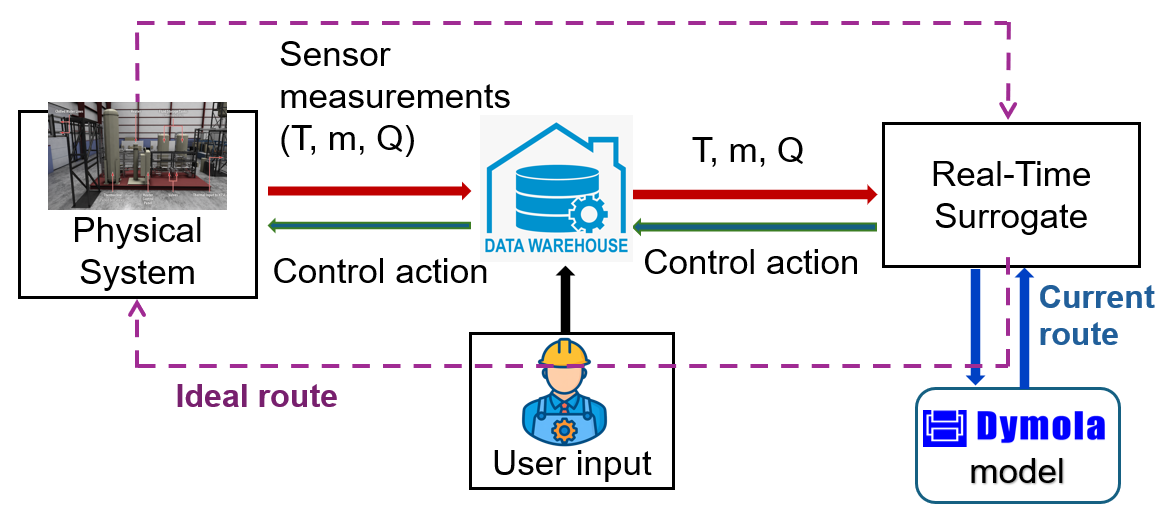}
    \caption{DT framework for autonomous control of TEDS. 
    The ideal route (purple) corresponds to direct training on experimental data, 
    whereas the current route (blue) uses simulation data 
    for surrogate training.}
    \label{fig:DT_framework}
\end{figure}

Integration of experimental data and physics-based simulations is achieved through a DT framework for supervisory control of TEDS. As illustrated in Figure~\ref{fig:DT_framework}, the DT connects the physical system with Modelica simulations, and real-time surrogate models.  

In the ideal route, surrogate models would be trained and updated directly from continuous experimental measurements of temperature, flow, and heat rate \((T, \dot{m}, Q)\), thereby closing the DT loop with physical data. However, as the experimental cost and availability constraints make this infeasible, the current route is adopted: physics-based simulations are used to generate synthetic data, which could encompass a wide range of potential experimental scenarios, and a surrogate model (FNN, GRU, or SINDyC-based) is then trained to emulate the system dynamics in real-time, enabling rapid predictions and UQ.  

\subsection{Sparse Identification of Nonlinear Dynamics with Control (SINDyC)}

The SINDyC framework discovers governing equations from time-series measurements in the presence of control inputs \cite{brunton2016discovering, kaptanoglu2021pysindy}. In brief, the time derivative of the state vector $\mathbf{x}$ is expressed as a sparse linear combination of candidate functions of states and inputs:

\begin{equation}
  \frac{d\mathbf{x}}{dt} \;\approx\; \Xi \,\Theta(\mathbf{x}, \mathbf{u}),
  \label{eq:sindyc_general}
\end{equation}
where $\Theta(\mathbf{x},\mathbf{u})$ is a library of basis functions and $\Xi$ is a coefficient matrix identified via sparsity-promoting regression (e.g., sequential thresholding or LASSO). This yields a parsimonious model suitable for forecasting and control.
 
In this work, we restricted the candidate library to linear terms in states and controls, thus producing linear state-space surrogates that are easy to interpret and integrate into the control system \cite{seurincontrol}. For TEDS---when using the states and actuators shown in Figure~\ref{fig:GHX}---the surrogates read:

\begin{equation}
\dot{\mathbf{x}}_{\mathrm{TES}}
= \mathbf{A}_{\mathrm{TES}}\mathbf{x}_{\mathrm{TES}}
+ \mathbf{B}_{\mathrm{TES}}\mathbf{u}_{\mathrm{TES}}
+ \mathbf{d}_{\mathrm{TES}},
\end{equation}
\begin{equation}
\dot{\mathbf{x}}_{\mathrm{GHX}}
= \mathbf{A}_{\mathrm{GHX}}\mathbf{x}_{\mathrm{GHX}}
+ \mathbf{B}_{\mathrm{GHX}}\mathbf{u}_{\mathrm{GHX}}
+ \mathbf{d}_{\mathrm{GHX}},
\end{equation}

with
\[
\mathbf{x}_{\mathrm{TES}} =
\big[\dot m_{\mathrm{TES,in}},\,
T_{\mathrm{TES,out}},\,
T_{\mathrm{top}},\,
T_{\mathrm{mid}},\,
T_{\mathrm{bot}} \big]^{\!\top},\quad
\mathbf{x}_{\mathrm{GHX}} =
\big[\dot m_{\mathrm{GHX}},\, Q_{\mathrm{GHX}} \big]^{\!\top},
\]
\[
\mathbf{u}_{\mathrm{TES}} = \mathbf{u}_{\mathrm{GHX}} =
\big[ PV_{006},\, \dot m_{\mathrm{pump,out}},\, T_{\mathrm{pump,in}},\, T_{\mathrm{heater,out}} \big]^{\!\top}.
\]
The matrices $\mathbf{A}$ and $\mathbf{B}$ are identified directly from the data. $\mathbf{d}_{\mathrm{TES}}\!\in\!\mathbb{R}^{5}$ and $\mathbf{d}_{\mathrm{GHX}}\!\in\!\mathbb{R}^{2}$ are learned constant offset (bias) vectors---the intercepts of each state equation arising from the constant (all-ones) column in the SINDyC library. They capture steady-state drift/unmodeled effects and sensor biases.

A single deterministic fit can be sensitive to noise and limited trajectories. To stabilize the surrogate and obtain calibrated uncertainty, we built a pool of linear SINDyC models in the following manner. From a total of 374 valid simulation trajectories, we randomly sampled 500 distinct subsets, each containing four trajectories. Each subset was used to train one linear SINDyC model, yielding 500 coefficient vectors. We then fit a multivariate Gaussian to these coefficient vectors, defining the MvG-SINDyC. This preserved interpretability (explicit linear state-space form) while providing the uncertainty needed for robust DT/MPC.

In an associated preliminary study, we repeated the same procedure with larger subset sizes (ranging from 4 to 16 trajectories per model) and evaluated the resulting error–coverage trade-offs \cite{nabila2026active}. For a fixed computational budget, the baseline configuration of 4 trajectories per model used in this work yielded the most favorable balance (a lower RMSE with well-calibrated 95\% coverage); consequently, we adopted this configuration throughout this paper.

\subsection{Machine Learning Surrogates: Feedforward Neural Network and Gated Recurrent Unit}
\label{sec:ml_models}

To complement the physics-informed SINDyC surrogates, two data-driven ML models were developed: an FNN and a GRU network. These models served as flexible approximators capable of capturing nonlinear relationships and temporal dependencies directly from the data, without requiring an explicit governing equation form.

The FNN is the simplest form of a deep learning surrogate, and consists of sequential layers of neurons that transform input vectors through nonlinear activation functions \cite{goodfellow2016deep}. Each hidden layer performs an affine transformation followed by a nonlinear mapping, enabling the network to approximate complex input-output relationships. For this study, the FNN was trained to predict the future GHX states---mass flow rate and heat transfer rate---given actuator signals as inputs. Its simplicity and differentiability made it computationally efficient and suitable for mapping TEDS behavior.

Two variants of the FNN were trained to assess the impact of experimental data on model generalization. The first, denoted as FNN (with EXP), incorporated both simulation and experimental data during training, providing the network with direct exposure to real system dynamics and noise characteristics. The second, FNN (w/o EXP), was trained exclusively on simulation-generated trajectories, representing a purely synthetic surrogate. This distinction, as will be shown later, helps us assess whether or not FNN has overfitting tendencies. 

The GRU extends this capability to temporal sequences, enabling the model to learn dynamic behaviors and dependencies over time. Unlike traditional recurrent neural networks, GRUs introduce gating mechanisms that regulate information flow between successive time steps, allowing the network to retain long-term dependencies while also mitigating vanishing or exploding gradient issues \cite{chung2014empirical}. The GRU’s hidden state acts as a compact memory of past observations, updated through reset and update gates that determine how much past information should be retained and/or overwritten. This structure enables GRUs to model the transient responses of the GHX and TES systems more effectively than can be done by purely feedforward architectures. Note that since AL requires repeated model retraining, the authors employed a GRU model instead of the more complex long short-term memory network. Although long short-term memory (LSTM) is a widely used recurrent architecture, the GRU proved sufficient in this study, affording a favorable balance between computational efficiency and predictive accuracy.

\subsection{Active Learning for Trajectory Sampling}
\label{sec:aal}

\begin{figure}[!h]
    \centering
    \includegraphics[width=0.9\linewidth]{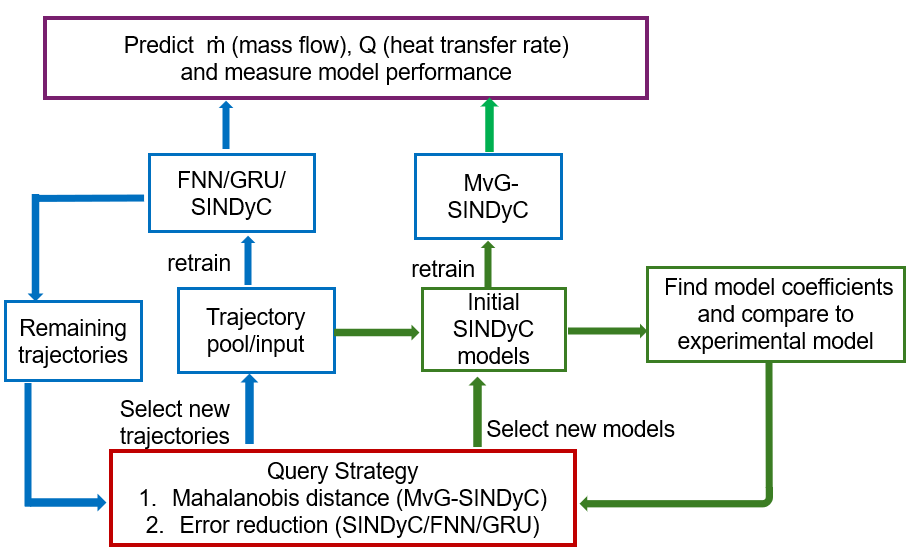}
    \caption{AL workflow for physics-informed (SINDyC/MvG-SINDyC) and data-driven (FNN/GRU) surrogates.}
    \label{fig:al_teds}
\end{figure}

The proposed methodology integrates AL into the development of surrogate models for the TEDS DT. Figure~\ref{fig:al_teds} illustrates the overall workflow, which consists of two complementary branches. Both branches rely on a unique AL query strategy that adaptively selects the most informative training trajectories.

The right-hand branch of the framework employs MvG-SINDyC to approximate the governing dynamics of TES-GHX states. Initially, from the pool of 374 valid simulation trajectories, 500 SINDyC models are constructed, where each model is trained using a small subset of four randomly selected trajectories. Each SINDyC model yields a coefficient vector defining a linear state-space surrogate consistent with Eqs. (6)–(7). The coefficients from all candidate models are then compared against those obtained from the experimentally identified SINDyC model to quantify structural discrepancy. AL iteratively identifies models that minimize the Mahalanobis distance between the simulation model and experiment model in the coefficient space. These selected models are then subsequently used to fit a multivariate Gaussian (MvG), which serves as the probabilistic surrogate representation of the TES–GHX dynamics. The loop proceeds until the surrogate achieves stable predictive performance.

\begin{figure}[!h]
    \centering
    \includegraphics[width=0.7\linewidth]{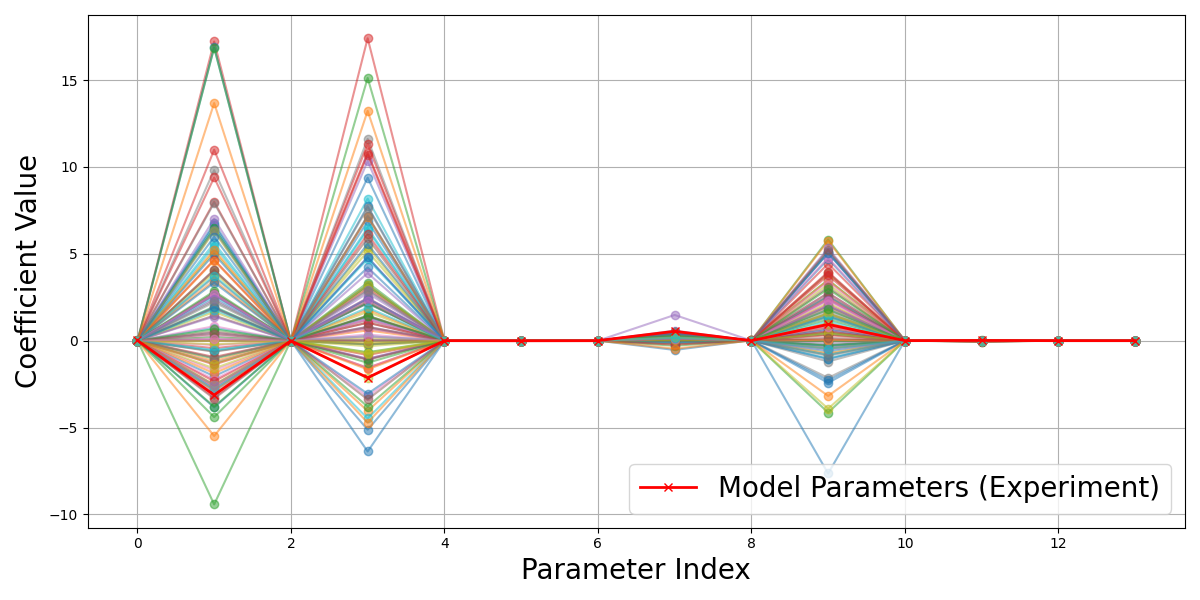}
    \caption{SINDyC model coefficient comparison. Thin lines: candidate models trained on individual sets of four simulation trajectories; bold red line: model coefficients based on experimental data.}
    \label{fig:mahal}
\end{figure}

Figure~\ref{fig:mahal} shows comparison among SINDyC model coefficient vectors from candidate simulation–trained models (thin curves) against the experimentally derived coefficient vector (bold red). The horizontal axis enumerates the SINDyC basis terms, and the vertical axis shows the associated coefficient values. Since the GHX state vector consists of two states, $\dot m_{\mathrm{GHX}}$ and $Q_{\mathrm{GHX}}$, and the library is restricted to linear terms, each state equation contains seven coefficients: one constant offset, two state-dependent terms, and four control-dependent terms. Consequently, the first seven coefficients correspond to the bypass mass-flow-rate dynamics, while the remaining seven correspond to the heat-transfer-rate dynamics, consistent with Eq. (7). Grouping coefficients in this manner enables a physically interpretable comparison of how each model represents mass-flow and thermal interactions.

To rank candidates, we measure how close a model’s coefficient vector $\mathbf{a}\in\mathbb{R}^P$ is to the experimental vector $\mathbf{a}_{\exp}\in\mathbb{R}^P$ using the Mahalanobis distance,

\begin{equation}
d_M(\mathbf{a},\mathbf{a}_{\exp})
=\sqrt{\bigl(\mathbf{a}-\mathbf{a}_{\exp}\bigr)^{\!\top}
\boldsymbol{\Sigma}^{-1}
\bigl(\mathbf{a}-\mathbf{a}_{\exp}\bigr)},
\label{eq:mahal}
\end{equation}
where $\boldsymbol{\Sigma}\in\mathbb{R}^{P\times P}$ is the empirical covariance matrix of the model coefficients, estimated from the current pool of candidates or selected models. This metric scales each direction by its variance and captures correlations among coefficients, so parameters that vary widely across models contribute less to the distance, while more stable, informative directions are emphasized. In the figure, the clustering of thin curves around the red line indicates that active selection based on \eqref{eq:mahal} drives the surrogate toward experimental dynamics with fewer training iterations. At each iteration, the MvG-SINDyC surrogate is updated and additional models are selected based on Mahalanobis distance in coefficient space. Although models with smaller distance are more consistent with the experimentally derived dynamics, a single closest model is insufficient to characterize uncertainty; instead, a set of nearby models is required to estimate a meaningful covariance structure. This selection balances fidelity to experimental behavior with sufficient diversity in parameter space to yield a well-conditioned probabilistic surrogate. As higher-distance models are added, their marginal contribution diminishes and may inflate uncertainty without improving predictive accuracy, naturally defining a practical cutoff. Compared with random sampling, the active learning strategy thus concentrates model selection in dynamically informative regions of coefficient space, enabling faster convergence of both accuracy and uncertainty calibration with fewer models.

The left-hand branch leverages black-box ML models (FNN and GRU) and deterministic SINDyC. A small set of initial trajectories is used to train each model. This model is then applied to predict the remaining trajectories. The active learning strategy is explicitly designed to identify trajectories where the model exhibits the largest prediction errors. These high-discrepancy cases are the most informative for improving model performance and are therefore prioritized for inclusion in subsequent training iterations. This iterative cycle improves predictive accuracy while also reducing reliance on large, randomly sampled datasets. At the core of the framework lies the query strategy, which unifies the two branches under a shared principle of information-driven sampling. In summary, two complementary query strategies are employed, depending on the surrogate model type:

\begin{itemize}
    \item For the MvG-SINDyC branch, the selection criterion is the Mahalanobis distance in model coefficient space. 
    \item For the SINDyC/FNN/GRU branch, the selection criterion is model prediction error.
\end{itemize}

For MvG-SINDyC, we also tested an error-based sampling strategy analogous to the approach used for the neural surrogates. However, this method did not yield measurable improvements in terms of convergence or robustness. Because the MvG-SINDyC framework represents model variability through the mean and covariance of the identified coefficient distributions, its informativeness is governed by structural diversity in parameter space, rather than pointwise prediction error. Consequently, error-based sampling provided no additional benefit, as the output residuals mainly reflected local noise rather than new dynamical information. In contrast, the Mahalanobis-distance criterion---defined in the coefficient space---proved more effective for identifying distinct dynamical regimes and was hence adopted as the default query strategy for the MvG-SINDyC branch. By focusing on those trajectories that contribute maximal information, AL ensures that both interpretable and black-box surrogates converge faster, even with fewer simulation or experimental runs.

\section{Results and Discussion}
\label{sec:results}

\begin{table}[!t]
\centering
\caption{Model architectures and training hyperparameters used in this study.}
\label{tab:hyperparams}
\begin{tabular}{p{3.0cm} p{6.4cm} p{6.8cm}}
\toprule
\textbf{Model type} & \textbf{Architecture} & \textbf{Hyperparameters} \\
\midrule
\textbf{SINDyC/ \newline MvG-SINDyC} 
& Linear candidate library (Polynomial Library, degree $=1$); optimizer: Sequential Thresholded Least Squares.
& $\lambda_{\text{TES}}=10^{-6}$, $\alpha_{\text{TES}}=10^{-3}$; $\lambda_{\text{GHX}}=10^{-8}$, $\alpha_{\text{GHX}}=10^{-6}$; integration via LSODA (rtol $=10^{-12}$, atol $=10^{-12}$). \\
\addlinespace[2pt]
\midrule
\textbf{FNN} 
& Feedforward network with four layers: $4 \rightarrow 128 \rightarrow 128 \rightarrow 2$ nodes with ReLU between linear layers. 
& Optimizer: Adam; learning rate $=10^{-3}$; loss: MSE; epochs $=40$.  \\
\midrule
\textbf{GRU} 
& GRU network: $6$ (features: past states + controls) $\rightarrow 128 \rightarrow 128 \rightarrow 2$ nodes; lookback $=120$; ReLU applied in output layer.
& Optimizer: Adam; learning rate $=10^{-3}$; loss: MSE; epochs $=20$. \\
\bottomrule
\end{tabular}
\end{table}

The architectures and training hyperparameters for each surrogate model---SINDyC, FNN, and GRU---are summarized in Table~\ref{tab:hyperparams}. The SINDyC models employed the Sequential Thresholded Least Squares algorithm---where the threshold parameter $\lambda$ controlled sparsity by eliminating small coefficients, and the ridge coefficient $\alpha$ provided L2 regularization to stabilize the least-squares fit. The reported $\mathrm{rtol}$ and $\mathrm{atol}$ values denote the relative and absolute tolerances of the LSODA integrator used for the model rollout. In the neural surrogates, the FNN followed a fully connected architecture with ReLU activations, and was optimized using the Adam algorithm, with a learning rate of $10^{-3}$. FNN (with EXP) and FNN (w/o EXP) shared the same architecture, and differed only in regard to whether experimental data were included during training. The GRU network processed temporal sequences by using a lookback window of $120$ time steps, with each input vector being comprised of past state variables and control signals. This lookback size defined the temporal context available to the model. The value of $120$ was selected based on a systematic study that evaluated prediction accuracy and computational cost across multiple sequence lengths (20–360 steps). Among the tested configurations, a lookback of 120 time steps provided a balanced trade-off: it achieved the lowest RMSE for both mass-flow and heat-transfer outputs while maintaining moderate training and inference times. All the models were trained with mean-squared-error loss and normalized input-output features to ensure stable convergence and consistent comparison across methods.

\begin{figure}[!h]
    \centering
    \includegraphics[width=0.8\linewidth]{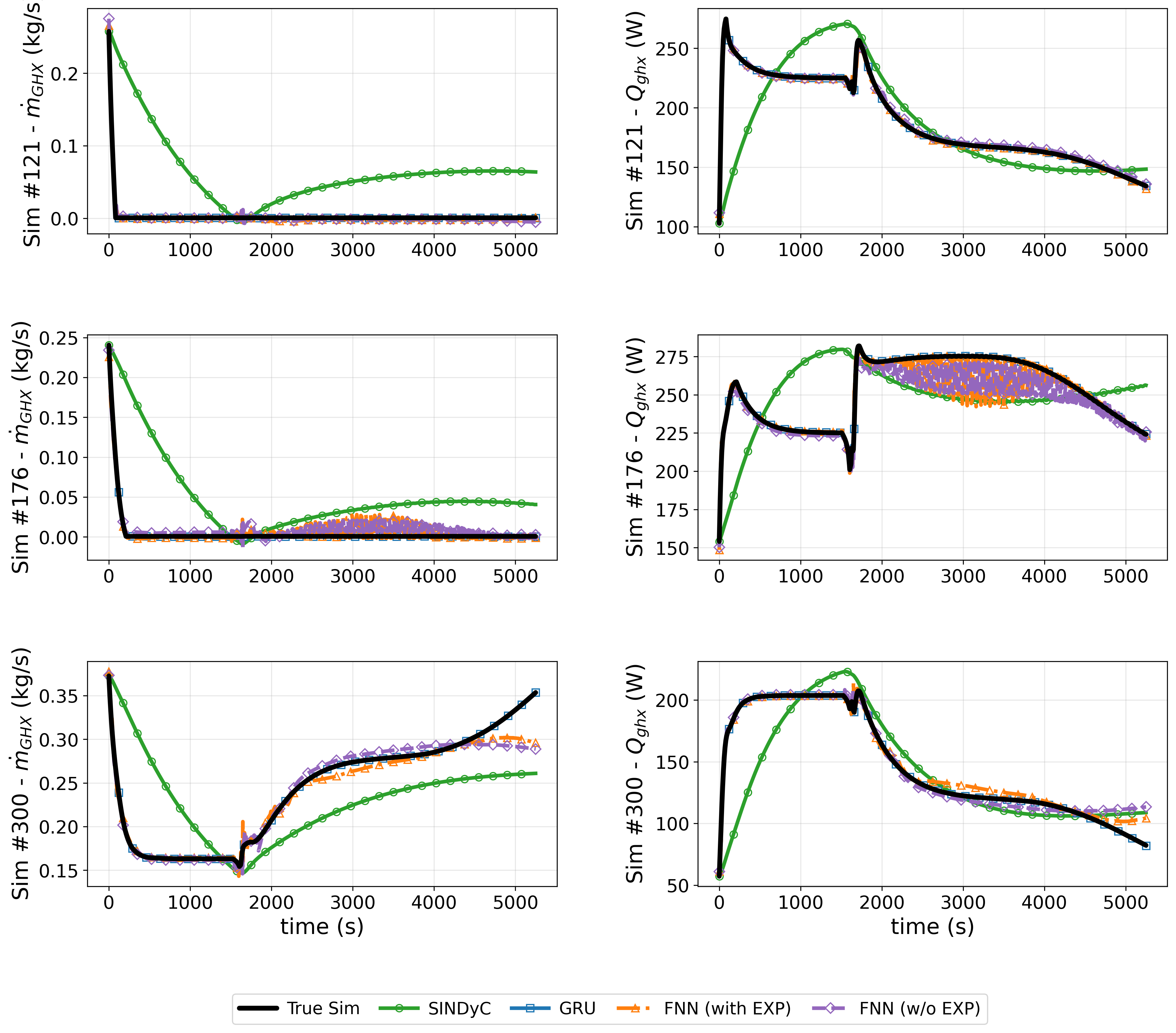}
    \caption{Comparison of the surrogate model predictions for three unseen simulation trajectories.}
    \label{fig:sim_pred}
\end{figure}

\begin{figure}[!h]
    \centering
    \includegraphics[width=0.8\linewidth]{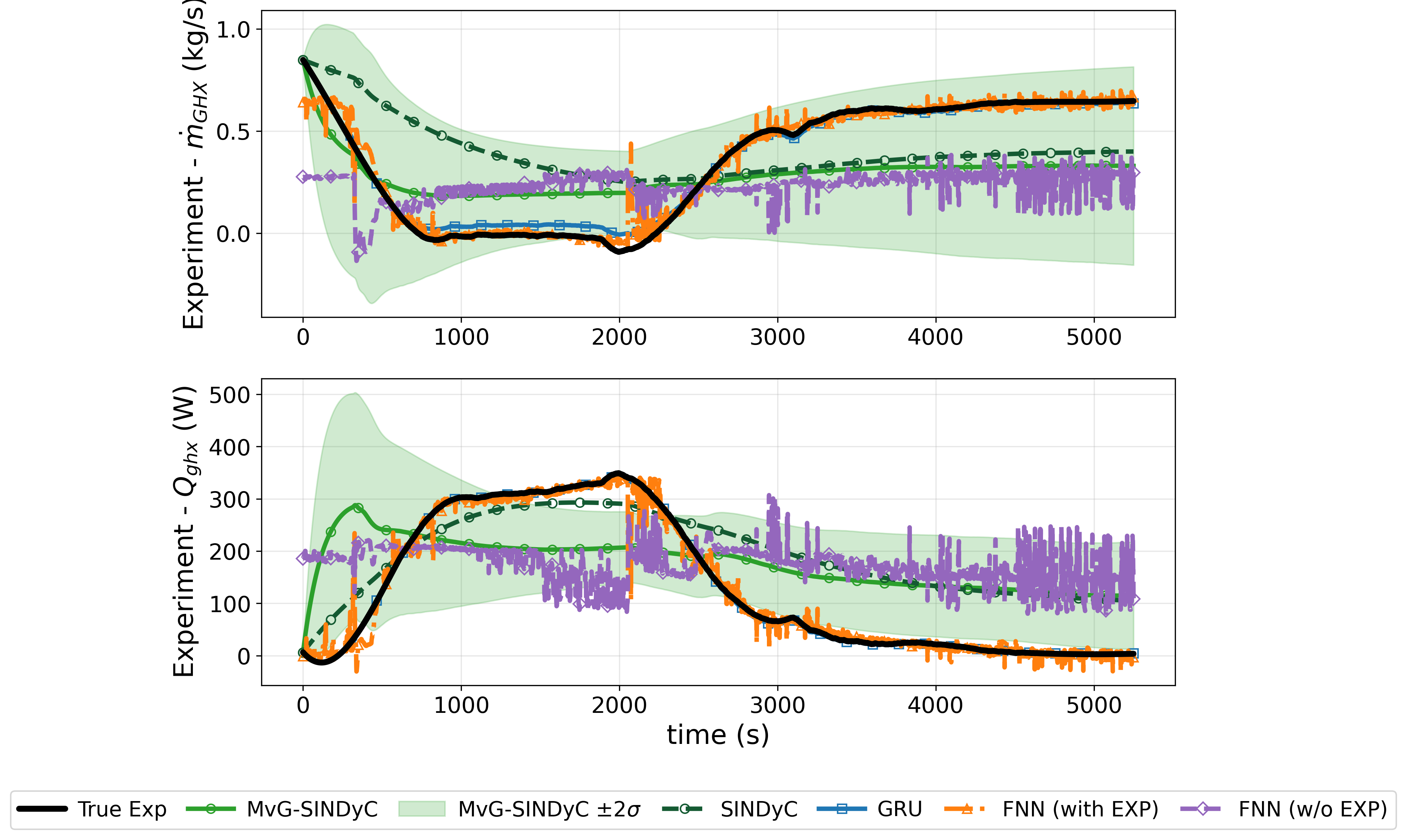}
    \caption{Comparison of surrogate model for experiment trajectory prediction.}
    \label{fig:exp_pred}
\end{figure}

Figs.~\ref{fig:sim_pred} and~\ref{fig:exp_pred} compare the surrogate model predictions against three unseen simulation trajectories and the experimental data, respectively.
In the simulation case (Figure~\ref{fig:sim_pred}), the GRU and FNN surrogates reproduce the reference trajectory with high fidelity, whereas SINDyC fails to capture the nonlinear transitions. 
This outcome highlights the trade-off between interpretability and accuracy: although SINDyC provides an analytical linear surrogate, it cannot resolve complex nonlinear behaviors inherent in the simulation data. Also, it is worth mentioning that GRU generally shows better predictive performance than FNN in Figure~\ref{fig:sim_pred}, taking advantage of its temporal features. 

In the experimental case (Figure~\ref{fig:exp_pred}), the FNN trained with experimental supervision and GRU achieve the best alignment with the measured GHX mass flow and heat transfer, accurately capturing the temporal dynamics. 
The FNN trained without experimental guidance shows larger variance and deviation, particularly in the steady-state region, underscoring the importance of experimental calibration.  Unlike GRU, FNN is a memoryless model that relies solely on instantaneous inputs and therefore cannot explicitly account for temporal history. As a result, it is more sensitive to simulation–experiment mismatch and requires experimental data to maintain accuracy. In contrast, the GRU incorporates a lookback window that enables it to capture GHX transient behavior more effectively, resulting in improved robustness without experimental supervision. In contrast, the deterministic SINDyC surrogate systematically underestimates the long-term dynamics owing to its linear coefficient formulation, which limits its capacity to represent strongly nonlinear transient–steady interactions. Note that in this case, the FNN model generally fails to generalize beyond the simulation data, exhibiting clear signs of overfitting and poor performance as compared to the experimental results. The proposed MvG-SINDyC framework extends the classical deterministic SINDyC approach by incorporating a probabilistic multivariate Gaussian representation, yielding both the mean trajectory and a time-varying uncertainty band. As shown in Figure~\ref{fig:exp_pred}, the mean prediction captures the overall experimental trend while the shaded 95\% credible interval realistically bounds most of the experimental data. The predictive uncertainty of the MvG-SINDyC model was further quantified by evaluating the fraction of the experimental trajectory lying within its $95\%$ predictive interval. This metric measures the uncertainty calibration of the probabilistic surrogate, i.e., the consistency between predicted credible regions and observed experimental variability. The model achieved coverage values of $0.89$ for the GHX mass-flow rate and $0.96$ for the heat-transfer rate. 

\begin{table}[!t]
\centering
\caption{Prediction error (RMSE) for three representative unseen simulations and the experiment, for $\dot{m}_{\mathrm{GHX}}$ and $Q_{\mathrm{GHX}}$.}
\label{tab:rmse_sim_exp}
\renewcommand{\arraystretch}{1.2}
\begin{tabular}{lcccccccc}
\toprule
 & \multicolumn{6}{c}{\textbf{Simulation RMSE}} & \multicolumn{2}{c}{\textbf{Experiment RMSE}} \\
\cmidrule(lr){2-7} \cmidrule(lr){8-9}
\textbf{Model} & 
\multicolumn{2}{c}{Sim \#121} & 
\multicolumn{2}{c}{Sim \#176} & 
\multicolumn{2}{c}{Sim \#300} & 
\multicolumn{2}{c}{(Exp.)} \\
\cmidrule(lr){2-3} \cmidrule(lr){4-5} \cmidrule(lr){6-7} \cmidrule(lr){8-9}
 & $\dot{m}$ (kg/s) & $Q$ (W) & $\dot{m}$ (kg/s) & $Q$ (W) & $\dot{m}$ (kg/s) & $Q$ (W) & $\dot{m}$ (kg/s) & $Q$ (W) \\
\midrule
\textbf{SINDyC} & 0.052 & 25.055 & 0.064 & 31.344 & 0.061 & 28.001 & 0.291 & 98.673 \\
\textbf{MvG-SINDyC} & --- & --- & --- & --- & --- & --- & 0.257 & 121.471 \\
\textbf{GRU} & 0.000 & 0.118 & 0.000 & 0.311 & 0.000 & 0.094 & 0.033 & 2.630 \\
\textbf{FNN (with EXP)} & 0.003 & 1.595 & 0.009 & 10.153 & 0.012 & 5.474 & 0.147 & 65.005 \\
\textbf{FNN (w/o EXP)} & 0.003 & 2.159 & 0.008 & 8.730 & 0.015 & 7.511 & 0.288 & 133.847 \\
\bottomrule
\end{tabular}
\end{table}

Table~\ref{tab:rmse_sim_exp} provides quantitative support to these observations, reporting the RMSEs for the three representative simulation trajectories (i.e., 121, 176, and 300) as well as the experimental trajectory. Across all simulations, the \textsc{GRU} is consistently the most accurate: the \(\dot m\) errors are essentially zero and the \(Q\) errors remain small (about \(0.094\text{--}0.311~\mathrm{W}\)). The FNN is competitive but less precise. With experimental supervision, the \textsc{FNN} attains an \(\dot m\) RMSE of \(0.003\text{--}0.012~\mathrm{kg/s}\) and a \(Q\) RMSE of \(1.595\text{--}10.153~\mathrm{W}\). Without supervision, it yields \(0.003\text{--}0.015~\mathrm{kg/s}\) and \(2.159\text{--}8.730~\mathrm{W}\). In contrast, the linear \textsc{SINDyC} surrogate is roughly an order of magnitude less accurate on these nonlinear trajectories (e.g., \(\dot m\) RMSE \(0.052\text{--}0.064~\mathrm{kg/s}\) and \(Q\) RMSE \(25\text{--}31~\mathrm{W}\)), highlighting that linear surrogates such as SINDyC are less suited for capturing nonlinear simulation dynamics.  

In the experimental case, performance differences are amplified. GRU remains the most accurate, with RMSE values of \(0.033~\mathrm{kg/s}\) and \(2.630~\mathrm{W}\), demonstrating strong generalization from simulation-trained models to physical system data. The FNN with experimental supervision improves significantly over SINDyC, but still exhibits moderate errors (\(0.147~\mathrm{kg/s}\), \(65.005~\mathrm{W}\)). 
The FNN without experimental data, however, performs the worst of all, with an RMSE exceeding that of SINDyC and MvG-SINDyC, reinforcing the necessity of experimental calibration. SINDyC models offer physically interpretable surrogates but underperform in nonlinear regimes; GRU achieves the best accuracy and robustness, particularly in generalizing from simulations to experiments; and FNN occupies an intermediate position, with access to experimental data substantially improving performance.

\begin{figure}[!h]
    \centering
    \includegraphics[width=0.9\linewidth]{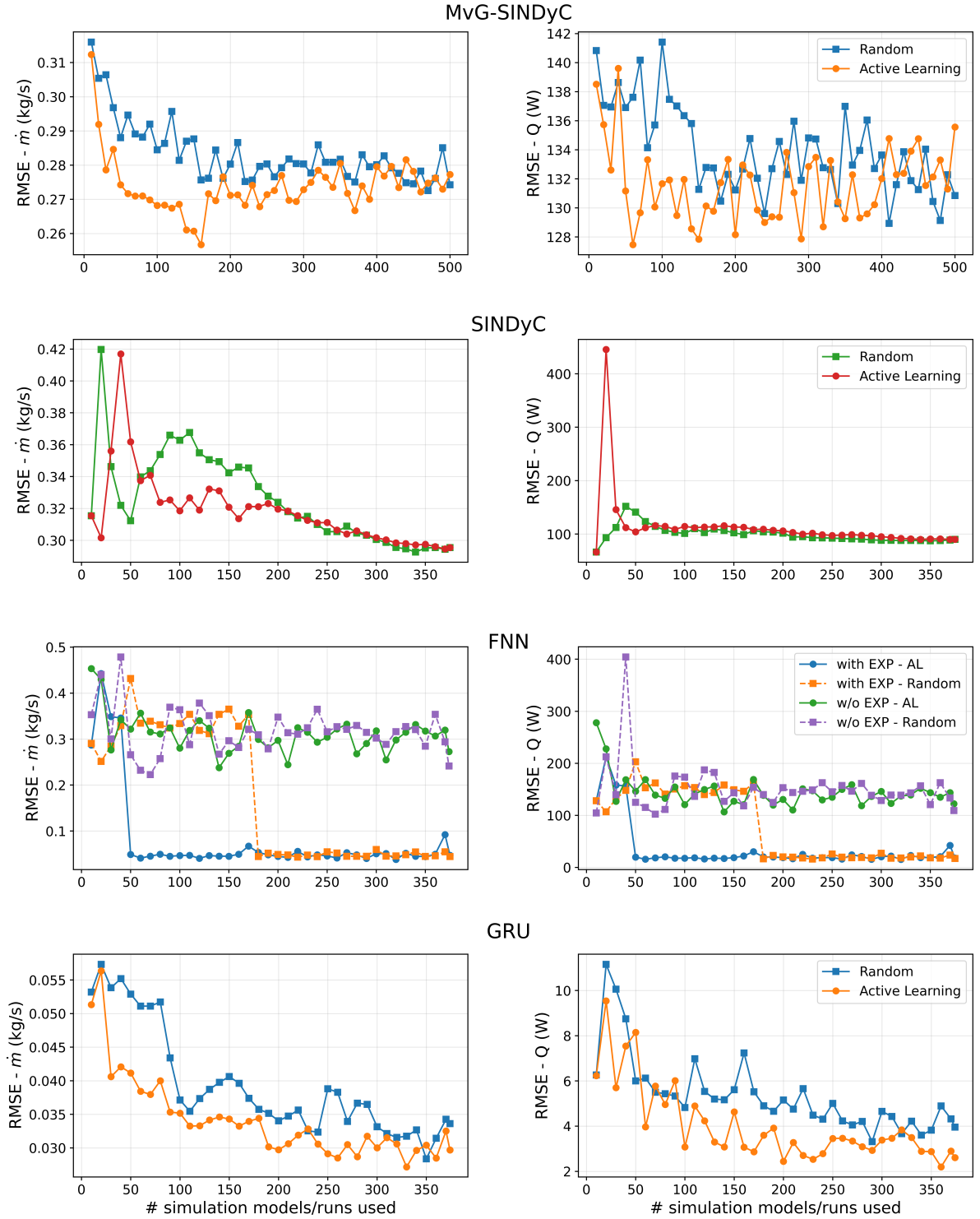}
    \caption{Prediction error trends (RMSE) pertaining to the mass-flow and power output for different models.}
    \label{fig:al_rmse}
\end{figure}

\begin{figure}[!h]
    \centering
    \includegraphics[width=0.8\linewidth]{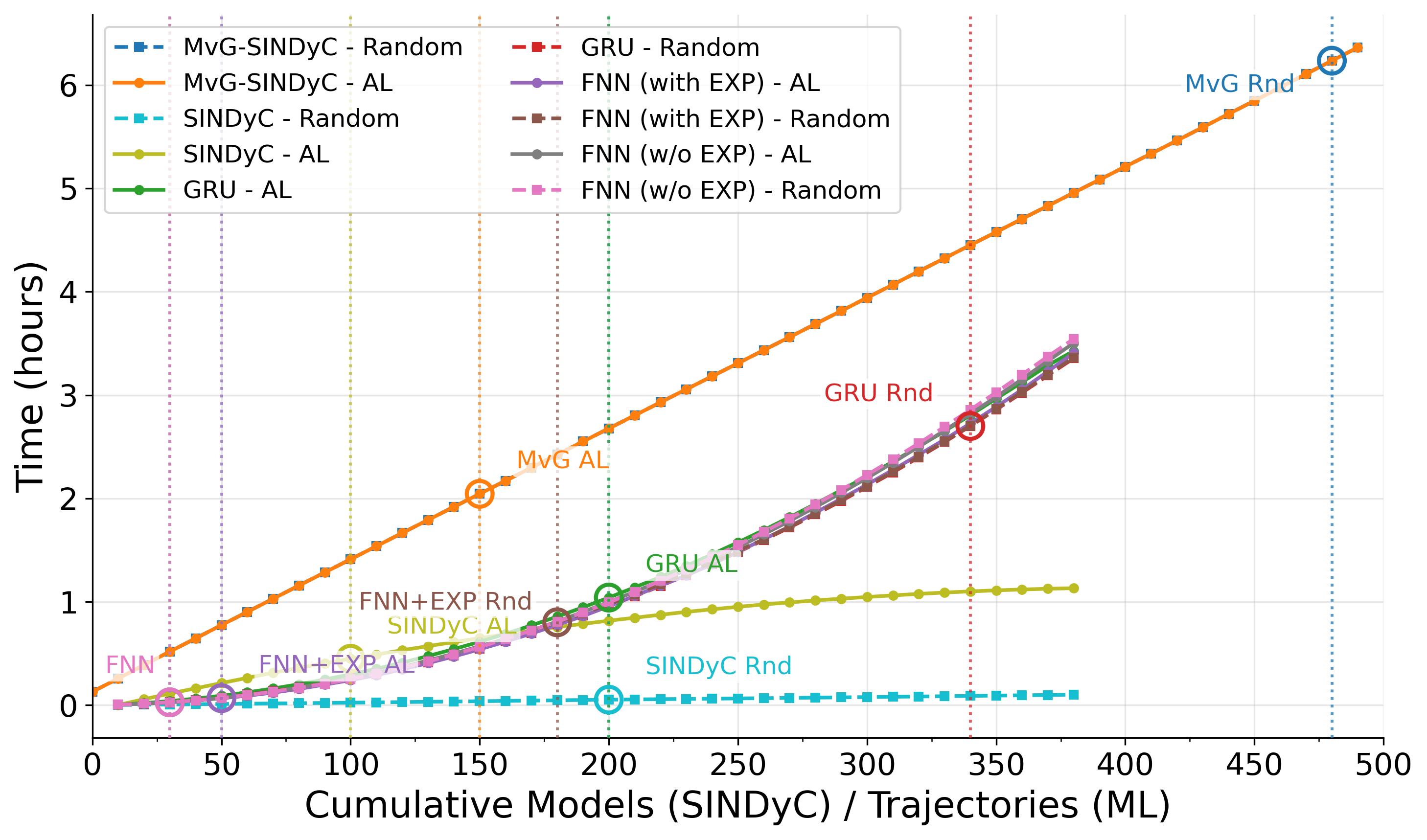}
    \caption{Computational time comparison of each model.}
    \label{fig:comp_time}
\end{figure}

Figure~\ref{fig:al_rmse} compares the prediction error trends of the four surrogate models (SINDyC, MvG-SINDyC,  FNN, GRU) under both AL and random sampling. 

For MvG-SINDyC (top row), AL consistently outperforms random sampling by producing a smoother convergence and lower RMSE for both quantities. The early-stage advantage (50--150 models) is particularly clear, with AL reducing the RMSE of $\dot{m}_{\mathrm{GHX}}$ from approximately $0.32$ to $0.25~\mathrm{kg/s}$, and that of $Q_{\mathrm{GHX}}$ from about $140$ to $125~\mathrm{W}$. Although the long-run difference narrows as the number of models increases, AL achieves the same final accuracy with roughly half the data, thus confirming its benefit in terms of data efficiency. The observed increase in RMSE beyond approximately 175 trajectories reflects a saturation effect rather than a limitation of the active learning (AL) strategy. AL prioritizes the most informative trajectories early in the process; once the dominant system dynamics are captured, adding additional trajectories --often redundant or weakly informative --does not necessarily improve, and may slightly degrade performance due to increased model variance and finite retraining noise. In this strategy, the AL framework is used to identify this cutoff strategically, enabling the training process to be terminated once performance gains plateau. Therefore, the increase in RMSE at larger training sizes does not indicate a lack of foresight in AL, but rather highlights its ability to efficiently determine an appropriate stopping point for data acquisition. This result reinforces the primary advantage of AL: achieving comparable accuracy with substantially fewer training trajectories, rather than slow improved performance at large data volumes.

For the deterministic SINDyC (second row), both AL and random sampling exhibit nearly identical convergence behavior. Because this linear surrogate cannot capture the strongly nonlinear dynamics of the GHX subsystem, the overall RMSE remains flat and relatively high, similar to the FNN trained without experimental data. Although AL employed an error-based sampling strategy analogous to that used for the neural surrogates, its impact was minimal—reflecting the limited information gain. The RMSE of $\dot{m}_{\mathrm{GHX}}$ stabilizes around $0.30~\mathrm{kg/s}$, and that of $Q_{\mathrm{GHX}}$ around $100~\mathrm{W}$, with both metrics showing only marginal reduction under AL. These results highlight that, for purely deterministic and linearized formulations such as SINDyC, the benefits of AL are inherently limited by model capacity rather than data coverage.

For the FNN surrogates (third row), AL yields the most dramatic improvement. When experimental data are included during training, the RMSE for $\dot{m}_{\mathrm{GHX}}$ drops from $\sim0.3$ to below $0.02~\mathrm{kg/s}$, and that for $Q_{\mathrm{GHX}}$ drops from $\sim150$ to nearly $10~\mathrm{W}$ after only $\sim200$ trajectories, whereas random sampling remains higher and noisier. In contrast, the FNN without experimental conditioning stays at around $0.32~\mathrm{kg/s}$ and $150~\mathrm{W}$, showing that exposure to real-system signals is crucial for stable black-box learning. As shown in Figure~\ref{fig:al_rmse} (middle row), both models---FNN with EXP-AL, and FNN with EXP-Random---exhibit a sharp reduction in error after approximately 50 and 175 iterations, respectively. These points correspond to when the AL process requested that the experimental trajectory be sampled and incorporated into the training dataset, leading to significant improvement in model performance.  

The GRU surrogates (bottom row) also benefit from AL, showing a consistently lower RMSE than for random sampling, across all stages. AL drives $\dot{m}_{\mathrm{GHX}}$ below $0.03~\mathrm{kg/s}$ and $Q_{\mathrm{GHX}}$ below $2.5~\mathrm{W}$ within the first 100--150 models, whereas random sampling converges more slowly. Notably, GRUs converge more quickly than SINDyC and with less noise than FNNs, reflecting their strength in capturing sequential dependencies in time-series data. This suggests that GRUs are particularly well-suited for dynamical systems (e.g., TEDS) when sufficient simulation trajectories are available.

When compared with the runtime analysis in Figure~\ref{fig:comp_time}, these results reveal a clear trade-off between model interpretability and computational efficiency. The MvG-SINDyC models required the longest cumulative training time, reaching over $6~\mathrm{hours}$ for 500 models under random sampling. The training time reduces to approximately $2~\mathrm{hours}$ when guided by AL, representing a near threefold reduction in total runtime. This cost stems from the repeated sparse regression and integration steps in each model fit, which scale linearly with the number of trajectories. Despite the higher cost, the interpretability of MvG-SINDyC makes this gain from AL particularly valuable, as fewer but more informative models can be used to achieve comparable accuracy. 
On the other hand, the neural surrogates (FNN and GRU) trained substantially faster, with cumulative training times on the order of a few hours. These fast gains are enabled by GPU-parallelized batch optimization and simple loss evaluation, in contrast to the repeated regression and aggregation steps required by MvG-SINDyC. For the GRU, active learning reduced the total training time from approximately $3.0~\mathrm{hours}$ under random sampling to $1.1~\mathrm{hours}$ with AL. The FNN trained with experimental supervision converged even more rapidly, requiring about $0.9~\mathrm{hours}$ with random sampling and less than $0.2~\mathrm{hours}$ under AL. This particularly large reduction is observed due to FNN's rapid saturation of performance once the most informative trajectories are identified. By comparison, deterministic SINDyC fits remain computationally inexpensive, requiring well under $0.2~\mathrm{hours}$ for approximately 200 models under random sampling and about $1.1~\mathrm{hours}$ for the AL path.

Interestingly, while the total wall-clock time of neural surrogates is lower, MvG-SINDyC exhibits the most substantial absolute and structural gains from AL by eliminating redundant model fits. AL thus plays a dual role: reducing the number of training trajectories needed, and lowering the computational cost by avoiding uninformative samples. Overall, these comparisons demonstrate that AL yields measurable efficiency gains across all model classes, with the largest absolute savings being observed for computationally intensive probabilistic MvG-SINDyC model. In addition, consistent---though smaller---reductions were seen for data-driven neural surrogates. Together, Figures~\ref{fig:al_rmse} and~\ref{fig:comp_time} demonstrate that AL accelerates convergence in both error and runtime dimensions, enabling scalable surrogate development for TEDS DTs.

\section{Conclusions}
\label{sec:conc}

This study introduced an AL framework for constructing a physics-informed DT of TEDS, integrating interpretable and data-driven surrogate modeling approaches. Four surrogate architectures were evaluated—deterministic SINDyC, probabilistic MvG-SINDyC, FNN, and GRU---to explore trade-offs among interpretability, data efficiency, predictive accuracy, and computational cost. By combining physics-based, system-level Modelica simulations with experimental measurements, the proposed workflow enables automated model refinement, uncertainty quantification, and systematic cross-validation across physics-informed and machine-learning paradigms.

Among the evaluated surrogates, the GRU model achieved the lowest prediction errors and strongest temporal generalization, with simulation RMSE values of $0.094$--$0.311~\mathrm{W}$ and an experimental RMSE of $2.63~\mathrm{W}$ for power output. Similar performance was observed for mass flow rate. The FNN trained on experimental data achieved comparable accuracy despite its lighter architecture, making it attractive for real-time deployment. In contrast, the deterministic SINDyC captured basic linearized dynamics with minimal computational overhead, making it the lightest and fastest to train. However, due to its limited representational capacity and reliance on error-based sampling, AL offered only marginal gains in predictive accuracy. The probabilistic MvG-SINDyC incorporated multivariate Gaussian inference to quantify uncertainty, producing credible intervals that captured experimental behavior and improving stability through sampling in coefficient space.

AL consistently outperformed random sampling across all surrogate families, improving both prediction accuracy and computational efficiency. Across the MvG-SINDyC, FNN, and GRU models, AL achieved up to an order-of-magnitude gain in data efficiency and significant reduction in total training time. The runtime analysis revealed a complementary trade-off between interpretability and computational efficiency. MvG-SINDyC incurred the highest cumulative cost, requiring approximately $6~\mathrm{hours}$ to train 500 models under random sampling. This was reduced to about $2~\mathrm{hours}$ with AL---a threefold speedup. Neural surrogates were trained substantially faster due to simpler architecture. These results confirm that AL not only reduces the number of required trajectories but also shortens the total compute time by avoiding redundant or low-information samples. The largest relative gains were observed for the computationally intensive SINDyC model, while neural networks benefited from smaller-yet-consistent runtime improvements.

Overall, the results highlight the complementary strengths of the four paradigms: SINDyC and MvG-SINDyC for interpretability and UQ, FNN for computational simplicity, and GRU for temporal fidelity. The proposed AL-DT framework unifies these advantages, providing a scalable pathway toward adaptive, trustworthy, physics-grounded DTs for complex thermal-hydraulic systems.

The next stage of this research will extend AL beyond model training to encompass the DT's entire supervisory control loop. This integration will enable autonomous supervisory control capable of balancing exploration (data acquisition) and exploitation (control performance) while also ensuring physical consistency and operational safety. By fusing physics-based interpretability with AL, the extended AL-DT framework represents a step toward self-improving, trustworthy AI for resilient and decarbonized energy infrastructures.

\section*{Data Availability}
\label{sec:avail}

All scripts and datasets used to reproduce all the results in this work are available in this public GitHub repository: \url{https://github.com/aims-umich/TEDS_DT_AL}

\section*{Acknowledgment}

This work was supported through Idaho National Laboratory (INL) Laboratory Directed Research and Development (LDRD) Program Award 24A1081-116FP, under U.S.~Department of Energy Idaho Operations Office Contract DE-AC07-05ID14517. The authors acknowledge the use of INL’s high-performance computing resources, which significantly contributed to the modeling and analysis efforts presented in this work. 

\section*{CRediT Author Statement}

\begin{itemize}
    \item \textbf{Umme Mahbuba Nabila}: Conceptualization, Methodology, Software, Validation, Formal Analysis, Visualization, Investigation, Data Curation, Writing - Original Draft. 
    \item \textbf{Paul Seurin}: Conceptualization, Methodology, Resources, Funding Acquisition, Supervision, Project Administration, Writing - Original Draft.
    \item \textbf{Linyu Lin}: Conceptualization, Methodology, Resources, Funding Acquisition, Supervision, Project Administration, Writing - Original Draft.
    \item \textbf{Majdi I. Radaideh}: Conceptualization, Methodology, Resources, Funding Acquisition, Supervision, Project Administration, Writing - Original Draft. 
\end{itemize}


\bibliographystyle{elsarticle-num}
\setlength{\bibsep}{0pt plus 0.3ex}
{
\bibliography{references}}

@article{seurincontrol,
  title={Control Under Uncertainty for a Physics-Informed Model of a Thermal Energy Distribution System: Qualitative Analysis},
  author={Seurin, Paul and Lin, Linyu},
  journal={Available at SSRN 5667550}

}

@article{radaideh2025multistep,
  title={Multistep Criticality Search and Power Shaping in Nuclear Microreactors with Deep Reinforcement Learning},
  author={Radaideh, Majdi I and Tunkle, Leo and Price, Dean and Abdulraheem, Kamal and Lin, Linyu and Elias, Moutaz},
  journal={Nuclear Science and Engineering},
  pages={1--13},
  year={2025},
  publisher={Taylor \& Francis}
}

@techreport{qin2022thermal,
  title={Thermal Stress Modeling and Analysis of Packed-bed Thermocline Energy Storage Tank for INL Thermal Energy Distribution System (TEDS)},
  author={Qin, Sunming and Yoo, Jun Soo and Morton, Terry James},
  year={2022},
  institution={Idaho National Lab.(INL), Idaho Falls, ID (United States)}
}

@techreport{morton2020thermal,
  title={Thermal Energy Distribution System (TEDS) Startup},
  author={Morton, Terry James},
  year={2020},
  institution={Idaho National Laboratory (INL), Idaho Falls, ID (United States)}
}

@article{seurin2026uncertainty,
  title={Uncertainty quantification of a physics-informed model based on sparse identification of a Thermal Energy Distribution System},
  author={Seurin, Paul and Lin, Linyu},
  journal={Annals of Nuclear Energy},
  volume={226},
  pages={111865},
  year={2026},
  publisher={Elsevier}
}

@techreport{frick2020development,
  title={Development of the inl thermal energy distribution system (teds) in the modelica eco-system for validation and verification},
  author={Frick, Konor L and Bragg-Sitton, Shannon M and Rabiti, Cristian},
  year={2020},
  institution={Idaho National Laboratory (INL), Idaho Falls, ID (United States)}
}

@article{kouvaritakis2016model,
  title={Model predictive control},
  author={Kouvaritakis, Basil and Cannon, Mark},
  journal={Switzerland: Springer International Publishing},
  volume={38},
  number={13-56},
  pages={7},
  year={2016},
  publisher={Springer}
}

@techreport{frick2021validation,
  title={Validation and Verification for INL Modelica-based TEDS models Via Experimental Results},
  author={Frick, Konor L and Bragg-Sitton, Shannon M and Garrouste, Marisol},
  year={2021},
  institution={Idaho National Lab.(INL), Idaho Falls, ID (United States)}
}

@article{lin2024autonomous,
  title={Autonomous control for Heat-Pipe microreactor using Data-Driven model predictive control},
  author={Lin, Linyu and Oncken, Joseph and Agarwal, Vivek},
  journal={Annals of Nuclear Energy},
  volume={200},
  pages={110399},
  year={2024},
  publisher={Elsevier}
}

@article{brunton2016discovering,
  title={Discovering governing equations from data by sparse identification of nonlinear dynamical systems},
  author={Brunton, Steven L and Proctor, Joshua L and Kutz, J Nathan},
  journal={Proceedings of the national academy of sciences},
  volume={113},
  number={15},
  pages={3932--3937},
  year={2016},
  publisher={National Academy of Sciences}
}

@inproceedings{lin2024development,
  title={Development of Supervisory Control System for Thermal Energy Distribution System},
  author={Lin, Linyu},
  booktitle={2024 Pacific Basin Nuclear Conference, PBNC 2024},
  pages={435--444},
  year={2024},
  organization={American Nuclear Society}
}

@article{kaiser2018sparse,
  title={Sparse identification of nonlinear dynamics for model predictive control in the low-data limit},
  author={Kaiser, Eurika and Kutz, J Nathan and Brunton, Steven L},
  journal={Proceedings of the Royal Society A},
  volume={474},
  number={2219},
  pages={20180335},
  year={2018},
  publisher={The Royal Society Publishing}
}

@article{el2024nuclear,
  title={Nuclear and renewables in multipurpose integrated energy systems: A critical review},
  author={El-Emam, Rami S and Constantin, Alina and Bhattacharyya, Rupsha and Ishaq, Haris and Ricotti, Marco E},
  journal={Renewable and Sustainable Energy Reviews},
  volume={192},
  pages={114157},
  year={2024},
  publisher={Elsevier}
}

@article{arvanitidis2023nuclear,
  title={Nuclear-driven integrated energy systems: A state-of-the-art review},
  author={Arvanitidis, Athanasios Ioannis and Agarwal, Vivek and Alamaniotis, Miltiadis},
  journal={Energies},
  volume={16},
  number={11},
  pages={4293},
  year={2023},
  publisher={MDPI}
}

@article{bragg2020reimagining,
  title={Reimagining future energy systems: Overview of the US program to maximize energy utilization via integrated nuclear-renewable energy systems},
  author={Bragg-Sitton, Shannon M and Boardman, Richard and Rabiti, Cristian and O'Brien, James},
  journal={International Journal of Energy Research},
  volume={44},
  number={10},
  pages={8156--8169},
  year={2020},
  publisher={Wiley Online Library}
}

@article{gautam2025digital,
  title={Digital Real-Time Simulation and Power Quality Analysis of a Hydrogen-Generating Nuclear-Renewable Integrated Energy System},
  author={Gautam, Sushanta and Szczublewski, Austin and Fox, Aidan and Mahmud, Sadab and Javaid, Ahmad and Olowu, Temitayo O and Westover, Tyler and Khanna, Raghav},
  journal={Energies},
  volume={18},
  number={4},
  pages={937},
  year={2025},
  publisher={MDPI}
}

@article{williams2024modeling,
  title={Modeling and Optimization of a Nuclear Integrated Energy System for the Remote Microgrid on El Hierro},
  author={Williams, Logan and Doster, J Michael and Mikkelson, Daniel},
  journal={Energies},
  volume={17},
  number={23},
  pages={5826},
  year={2024},
  publisher={MDPI}
}

@article{hills2021dynamic,
  title={Dynamic modeling and simulation of nuclear hybrid energy systems using freeze desalination and reverse osmosis for clean water production},
  author={Hills, Stephen and Dana, Seth and Wang, Hailei},
  journal={Energy Conversion and Management},
  volume={247},
  pages={114724},
  year={2021},
  publisher={Elsevier}
}

@article{jacob2023modeling,
  title={Modeling and control of nuclear--renewable integrated energy systems: Dynamic system model for green electricity and hydrogen production},
  author={Jacob, Roshni Anna and Zhang, Jie},
  journal={Journal of Renewable and Sustainable Energy},
  volume={15},
  number={4},
  pages={046302},
  year={2023},
  publisher={AIP Publishing LLC}
}

@article{mikkelson2022analysis,
  title={Analysis of controls for integrated energy storage system in energy arbitrage configuration with concrete thermal energy storage},
  author={Mikkelson, Daniel and Frick, Konor},
  journal={Applied Energy},
  volume={313},
  pages={118800},
  year={2022},
  publisher={Elsevier}
}

@article{luxembourg2025times,
  title={TIMES-Europe: An integrated energy system model for analyzing Europe’s energy and climate challenges},
  author={Luxembourg, Stefan L and Salim, Steven S and Smekens, Koen and Longa, Francesco Dalla and van der Zwaan, Bob},
  journal={Environmental Modeling \& Assessment},
  volume={30},
  number={1},
  pages={1--19},
  year={2025},
  publisher={Springer}
}

@article{masotti2023modeling,
  title={Modeling and simulation of nuclear hybrid energy systems architectures},
  author={Masotti, Guido Carlo and Cammi, Antonio and Lorenzi, Stefano and Ricotti, Marco Enrico},
  journal={Energy Conversion and Management},
  volume={298},
  pages={117684},
  year={2023},
  publisher={Elsevier}
}

@article{de2000mahalanobis,
  title={The mahalanobis distance},
  author={De Maesschalck, Roy and Jouan-Rimbaud, Delphine and Massart, D{\'e}sir{\'e} L},
  journal={Chemometrics and intelligent laboratory systems},
  volume={50},
  number={1},
  pages={1--18},
  year={2000},
  publisher={Elsevier}
}

@article{lecun2015deep,
  title={Deep learning},
  author={LeCun, Yann and Bengio, Yoshua and Hinton, Geoffrey},
  journal={nature},
  volume={521},
  number={7553},
  pages={436--444},
  year={2015},
  publisher={Nature Publishing Group UK London}
}

@article{burnett2025variational,
  title={Variational Digital Twins},
  author={Burnett, Logan A and Nabila, Umme Mahbuba and Radaideh, Majdi I},
  journal={arXiv preprint arXiv:2507.01047},
  year={2025}
}

@conference{nabila2026active,
  Address   = {Torino, Italy, April 19--23},
  Author    = {Nabila, Umme M and Seurin, Paul and Lin, Linyu and Radaideh, Majdi I},
  Booktitle = {Proc. International Conference on the Physics of Reactors (PHYSOR 2026)},
  Title     = {Active Learning for Uncertainty Quantification of a Physics-Informed Digital Twin of a Thermal Energy Distribution System},
  Year      = {2026}
}

@article{chung2014empirical,
  title={Empirical evaluation of gated recurrent neural networks on sequence modeling},
  author={Chung, Junyoung and Gulcehre, Caglar and Cho, KyungHyun and Bengio, Yoshua},
  journal={arXiv preprint arXiv:1412.3555},
  year={2014}
}

@book{goodfellow2016deep,
  title={Deep learning},
  author={Goodfellow, Ian and Bengio, Yoshua and Courville, Aaron and Bengio, Yoshua},
  volume={1},
  number={2},
  year={2016},
  publisher={MIT press Cambridge}
}

@article{kaptanoglu2021pysindy,
  title={PySINDy: A comprehensive Python package for robust sparse system identification},
  author={Kaptanoglu, Alan A and de Silva, Brian M and Fasel, Urban and Kaheman, Kadierdan and Goldschmidt, Andy J and Callaham, Jared L and Delahunt, Charles B and Nicolaou, Zachary G and Champion, Kathleen and Loiseau, Jean-Christophe and others},
  journal={arXiv preprint arXiv:2111.08481},
  year={2021}
}

@article{radaideh2019combining,
  title={Combining simulations and data with deep learning and uncertainty quantification for advanced energy modeling},
  author={Radaideh, Majdi I and Kozlowski, Tomasz},
  journal={International Journal of Energy Research},
  volume={43},
  number={14},
  pages={7866--7890},
  year={2019},
  publisher={Wiley Online Library}
}

@article{saleem2020application,
  title={Application of deep neural networks for high-dimensional large BWR core neutronics},
  author={Saleem, Rabie Abu and Radaideh, Majdi I and Kozlowski, Tomasz},
  journal={Nuclear Engineering and Technology},
  volume={52},
  number={12},
  pages={2709--2716},
  year={2020},
  publisher={Elsevier}
}

@article{radaideh2020surrogate,
  title={Surrogate modeling of advanced computer simulations using deep Gaussian processes},
  author={Radaideh, Majdi I and Kozlowski, Tomasz},
  journal={Reliability Engineering \& System Safety},
  volume={195},
  pages={106731},
  year={2020},
  publisher={Elsevier}
}

@article{radaideh2020neural,
  title={Neural-based time series forecasting of loss of coolant accidents in nuclear power plants},
  author={Radaideh, Majdi I and Pigg, Connor and Kozlowski, Tomasz and Deng, Yujia and Qu, Annie},
  journal={Expert Systems with Applications},
  volume={160},
  pages={113699},
  year={2020},
  publisher={Pergamon}
}

@article{radaideh2022pesa,
  title={PESA: Prioritized experience replay for parallel hybrid evolutionary and swarm algorithms-Application to nuclear fuel},
  author={Radaideh, Majdi I and Shirvan, Koroush},
  journal={Nuclear Engineering and Technology},
  volume={54},
  number={10},
  pages={3864--3877},
  year={2022},
  publisher={Elsevier}
}

@article{radaideh2023neorl,
  title={NEORL: NeuroEvolution Optimization with Reinforcement Learning—Applications to carbon-free energy systems},
  author={Radaideh, Majdi I and Du, Katelin and Seurin, Paul and Seyler, Devin and Gu, Xubo and Wang, Haijia and Shirvan, Koroush},
  journal={Nuclear Engineering and Design},
  volume={412},
  pages={112423},
  year={2023},
  publisher={North-Holland}
}

@article{alanazi2023multi,
  title={Multi-module-based CVAE to predict HVCM faults in the SNS accelerator},
  author={Alanazi, Yasir and Schram, Malachi and Rajput, Kishansingh and Goldenberg, Steven and Vidyaratne, Lasitha and Pappas, Chris and Radaideh, Majdi I and Lu, Dan and Ramuhalli, Pradeep and Cousineau, Sarah},
  journal={Machine Learning with Applications},
  volume={13},
  pages={100484},
  year={2023},
  publisher={Elsevier}
}

@article{price2024simplified,
  title={Simplified matching pursuits applied to 3D nuclear reactor temperature distribution construction},
  author={Price, Dean and Radaideh, Majdi I and Kochunas, Brendan},
  journal={Applied Mathematical Modelling},
  volume={131},
  pages={134--158},
  year={2024},
  publisher={Elsevier}
}

@article{sene2025surrogate,
  title={Surrogate-driven Variance-based Sensitivity Analysis of Thermal Storage Tanks in Integrated Energy Systems},
  author={Sene, Seydou and Lin, Linyu and Kim, Junyung and Radaideh, Majdi I},
  journal={\textit{In: Nuclear Plant Instrumentation and Control \& Human-Machine Interface Technology (NPIC\&HMIT 2025)}, Chicago, Illinois, United States, June 15–18, 2025},
  year={2025},
  publisher={Authorea}
}

@article{tunkle2025nuclear,
  title={Nuclear microreactor transient and load-following control with deep reinforcement learning},
  author={Tunkle, Leo and Abdulraheem, Kamal and Lin, Linyu and Radaideh, Majdi I},
  journal={Energy Conversion and Management: X},
  pages={101090},
  year={2025},
  publisher={Elsevier}
}

@article{radaideh2021physics,
  title={Physics-informed reinforcement learning optimization of nuclear assembly design},
  author={Radaideh, Majdi I and Wolverton, Isaac and Joseph, Joshua and Tusar, James J and Otgonbaatar, Uuganbayar and Roy, Nicholas and Forget, Benoit and Shirvan, Koroush},
  journal={Nuclear Engineering and Design},
  volume={372},
  pages={110966},
  year={2021},
  publisher={Elsevier}
}

@inproceedings{radaideh2022application,
  title={Application of Convolutional and Feedforward Neural Networks for Fault Detection in Particle Accelerator Power Systems},
  author={Radaideh, Majdi and Pappas, Chris and Ramuhalli, Pradeep and Cousineau, Sarah},
  booktitle={Annual Conference of the PHM Society},
  volume={14},
  number={1},
  year={2022}
}

@article{radaideh2022time,
  title={Time series anomaly detection in power electronics signals with recurrent and ConvLSTM autoencoders},
  author={Radaideh, Majdi I and Pappas, Chris and Walden, Jared and Lu, Dan and Vidyaratne, Lasitha and Britton, Thomas and Rajput, Kishansingh and Schram, Malachi and Cousineau, Sarah},
  journal={Digital Signal Processing},
  volume={130},
  pages={103704},
  year={2022},
  publisher={Elsevier}
}

@article{radaideh2023early,
  title={Early fault detection in particle accelerator power electronics using ensemble learning},
  author={Radaideh, Majdi I and Pappas, Chris and Wezensky, Mark and Ramuhalli, Pradeep and Cousineau, Sarah},
  journal={International Journal of Prognostics and Health Management},
  volume={14},
  number={1},
  year={2023},
  publisher={Oak Ridge National Laboratory (ORNL), Oak Ridge, TN (United States)}
}

@article{li2023data,
  title={Data-driven distributionally robust scheduling of community integrated energy systems with uncertain renewable generations considering integrated demand response},
  author={Li, Yang and Han, Meng and Shahidehpour, Mohammad and Li, Jiazheng and Long, Chao},
  journal={Applied Energy},
  volume={335},
  pages={120749},
  year={2023},
  publisher={Elsevier}
}

@article{zhang2023hybrid,
  title={Hybrid data-driven method for low-carbon economic energy management strategy in electricity-gas coupled energy systems based on transformer network and deep reinforcement learning},
  author={Zhang, Bin and Hu, Weihao and Xu, Xiao and Zhang, Zhenyuan and Chen, Zhe},
  journal={Energy},
  volume={273},
  pages={127183},
  year={2023},
  publisher={Elsevier}
}

@article{zhou2025data,
  title={Data-driven distributionally robust stochastic optimal dispatching method of integrated energy system considering multiple uncertainties},
  author={Zhou, Yixing and Hou, Hongjuan and Yan, Haoran and Wang, Xi and Zhou, Rhonin},
  journal={Energy},
  volume={325},
  pages={136104},
  year={2025},
  publisher={Elsevier}
}

@article{prina2024machine,
  title={Machine learning as a surrogate model for EnergyPLAN: Speeding up energy system optimization at the country level},
  author={Prina, Matteo Giacomo and Dallapiccola, Mattia and Moser, David and Sparber, Wolfram},
  journal={Energy},
  volume={307},
  pages={132735},
  year={2024},
  publisher={Elsevier}
}

@article{ledee2025improved,
  title={Improved surrogate modeling for multi-energy system design: Model architecture, sampling and scaling choices},
  author={L{\'e}d{\'e}e, Fran{\c{c}}ois and Crawford, Curran and Evins, Ralph},
  journal={Applied Energy},
  volume={390},
  pages={125812},
  year={2025},
  publisher={Elsevier}
}

\end{document}